\documentclass[sn-basic, iicol]{sn-jnl}

\usepackage{graphicx}
\usepackage{multirow}%
\usepackage{amsmath,amssymb,amsfonts}%
\usepackage{amsthm}%
\usepackage{mathrsfs}%
\usepackage[title]{appendix}%
\usepackage[table,xcdraw]{xcolor}
\usepackage{textcomp}%
\usepackage{manyfoot}%
\usepackage{booktabs}%
\usepackage{algorithm}%
\usepackage{algorithmicx}%
\usepackage{algpseudocode}%
\usepackage{listings}%

\usepackage{makecell}
\usepackage{array}
\usepackage{booktabs}
\usepackage{pifont}  

\theoremstyle{thmstyleone}%
\theoremstyle{thmstyletwo}%

\theoremstyle{thmstylethree}%
\newcommand{\name}{AddSR}

\begin{document}

\title[Article Title]{AddSR: Accelerating Diffusion-based Blind Super-Resolution with Adversarial Diffusion Distillation}


\author[1]{\fnm{Rui} \sur{Xie}}\email{ruixie0097@gmail.com}

\author[1]{\fnm{Chen} \sur{Zhao}}\email{2518628273@qq.com}

\author[1]{\fnm{Kai} \sur{Zhang}}\email{kaizhang@nju.edu.cn}

\author[1]{\fnm{Zhenyu} \sur{Zhang}}\email{zhangjesse@foxmail.com}

\author[2]{\fnm{Jun} \sur{Zhou}}\email{zhouj@swu.edu.cn}

\author[1]{\fnm{Jian} \sur{Yang}}\email{csjyang@nju.edu.cn}

\author*[1]{\fnm{Ying} \sur{Tai}}\email{yingtai@nju.edu.cn}

\affil[1]{\orgname{State Key Laboratory for Novel Software Technology, Nanjing University}, \orgaddress{\state{Jiangsu}, \country{China}}}

\affil[2]{\orgname{Southwest University}, \orgaddress{\state{Chongqing}, \country{China}}}


\abstract{Blind super-resolution methods based on Stable Diffusion (SD) demonstrate impressive generative capabilities in reconstructing clear, high-resolution (HR) images with intricate details from low-resolution (LR) inputs. However, their practical applicability is often limited by poor efficiency, as they require hundreds to thousands of sampling steps.
Inspired by Adversarial Diffusion Distillation (ADD), we incorporate this approach to design a highly effective and efficient blind super-resolution method. Nonetheless, two challenges arise: First, the original ADD significantly reduces result fidelity, leading to a perception-distortion imbalance. Second, SD-based methods are sensitive to the quality of the conditioning input, while LR images often have complex degradation, which further hinders effectiveness.
To address these issues, we introduce a Timestep-Adaptive ADD (TA-ADD) to mitigate the perception-distortion imbalance caused by the original ADD. Furthermore, we propose a prediction-based self-refinement strategy to estimate HR, which allows for the provision of more high-frequency information without the need for additional modules.
Extensive experiments show that our method,~\name, generates superior restoration results while being significantly faster than previous SD-based state-of-the-art models (e.g., $7\times$ faster than SeeSR).}


\maketitle

\section{Introduction}\label{sec1}

\begin{figure*}[h]
    \centering
    \includegraphics[width=1\linewidth]{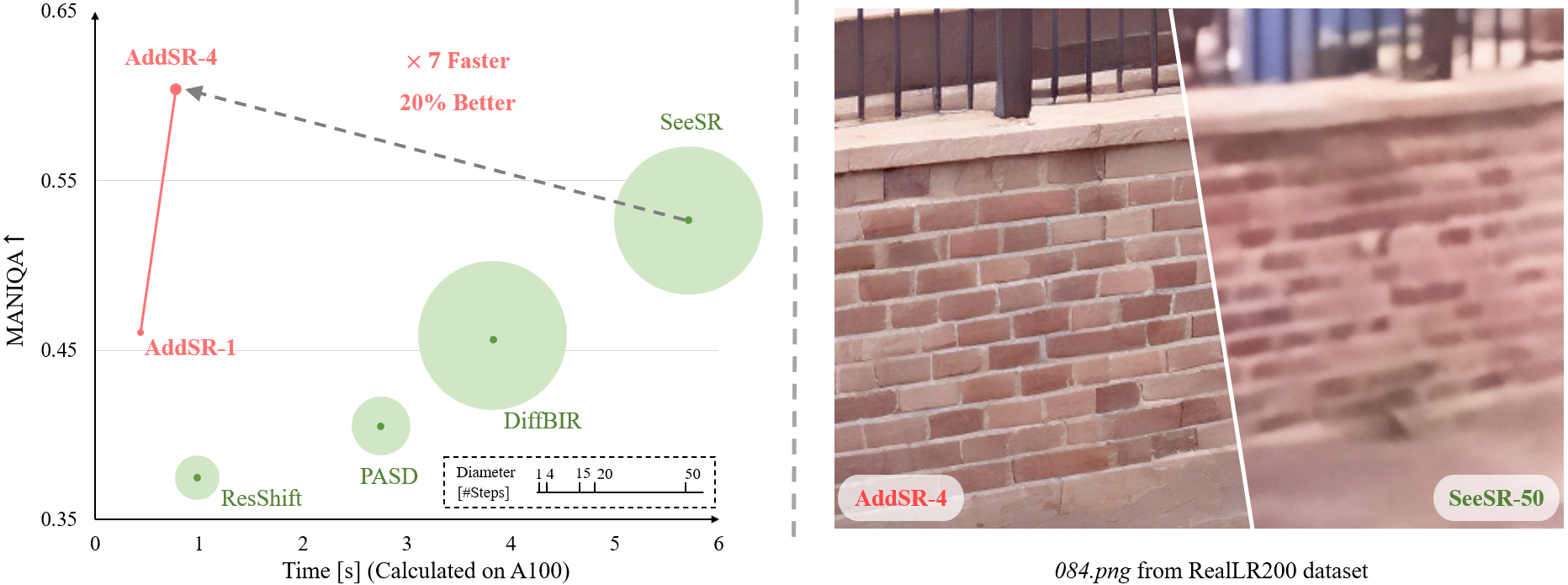}
    \caption{\textbf{Comparisons on effect and efficiency}.~\name-$4$ indicates the result is obtained in $4$ steps, achieving high perception quality restoration performance with the fastest speed among diffusion-based models. 
    In contrast, existing SD-based BSR models suffer from either low perception quality restoration performance (\textit{e.g.}, ResShift) or time-consuming efficiency (\textit{e.g.}, SeeSR-$50$).
    }
    \label{fig:introduction}
\end{figure*}

\label{sec:intro}
Blind super-restoration (BSR) aims to convert low-resolution (LR) images that have undergone complex and unknown degradation into clear high-resolution (HR) versions. Differing from classical super-resolution~\cite{SRCNN,VDSR,SRGAN,chen2021pre,zhang2022efficient,tai2017image,tai2017memnet}, where the degradation process is singular and known, BSR is crafted to enhance real-world degraded images, imbuing them with heightened practical value. 
 
Generative models, \textit{e.g.}~generative adversarial network (GAN) and diffusion model, have demonstrated significant superiority in BSR task to achieve realistic details. GAN-based~models~\cite{BSRGAN, realesrgan, KDSR, LDL, femasr, swinir} learn a mapping from the distribution of input LR images to that of HR images with adversarial training. 
However, when handling natural images with intricate textures, they often struggle to generate unsatisfactory visual results due to unstable adversarial objectives~\cite{LDL, xie2023desra}.

Recently, diffusion models (DM)~\cite{ddpm,ddim,nan2024openvid} have garnered significant attention owing to their potent generative capabilities and the ability to combine information from multiple modalities. 
DM-based BSR methods can be roughly divided into two categories: those without Stable Diffusion (SD) prior~\cite{yue2023resshift,SR3,2022SRDiff}, and those incorporating SD prior~\cite{DiffBIR,wu2023seesr,StableSR}. 
SD prior can significantly enhance the model's ability to capture the distribution of natural images~\cite{PASD}, thereby enabling the generated HR images with realistic details. 
Given the iterative refinement nature of DM, diffusion-based methods typically outperform GAN-based ones, albeit at the expense of efficiency. Hence, there's an urgent demand for BSR models that \textit{deliver exceptional restoration quality while maintaining high efficiency} for real-world applications.

To achieve the above goal, we draw inspiration from Adversarial Diffusion Distillation (ADD)~\cite{ADD} and introduce it into the BSR task. However, two key challenges still exist: 1) \textit{Perception-distortion imbalance}~\cite{perception, zhang2022perception, luo2024skipdiff}: Directly applying ADD in the BSR task leads to reduced fidelity, causing a perception-distortion imbalance that undermines effectiveness. 
2) \textit{Efficient restoration of high-frequency details}: The quality of the conditioning input can significantly affect the restored results \cite{liao2024denoisingadaptationnoisespacedomain}. Previous SD-based methods \cite{StableSR, DiffBIR} rely on additional degradation removal modules to pre-clean LR images for conditioning, which hinders efficiency. Therefore, efficiently obtaining a conditioning input with more high-frequency signals to guide restoration is a key challenge for effective and efficient blind super-resolution (BSR).

In this paper, we propose a novel AddSR based on ADD for blind super-restoration, which enhances restoration effects and accelerates inference speed of SD-based models simultaneously. 
There are two critical designs in AddSR to address the above issues respectively: 1) We introduce \textit{timestep-adaptive adversarial diffusion distillation (TA-ADD)} loss, which designs a bivariate timestep-related weighting function to achieve perception-distortion balance, enhancing generative ability at smaller inference steps while reducing it at larger ones. 2) We propose a simple yet effective strategy, \textit{prediction-based self-refinement (PSR)}, which uses the estimated HR image from the predicted noise to control the model output. This approach enables efficient condition restoration of the high-frequency components and further allows the restored results to contain more high-frequency details.
Our main contributions can be summarized as threefold:

$\bullet$ To the best of our knowledge, the proposed AddSR is the first to explore ADD for efficient and effective blind super-resolution, achieving a $\times$$7$ speedup over SeeSR~\cite{wu2023seesr} while delivering improved perceptual quality.

$\bullet$ We introduce a new TA-ADD loss to address the perception-distortion imbalance issue introduced by the original ADD, allowing AddSR to generate superior perceptual quality while maintaining comparable fidelity.

$\bullet$ We propose a prediction-based self-refinement strategy to efficiently restore condition and enable the restored results to generate more details without the need for additional modules. 

\section{Related Work}
\label{sec:related work}
\textbf{GAN-based BSR.}
In recent years, BSR have drawn much attention due to their practicability. 
Adversarial training~\cite{GAN, park2023content, AND, zhang2024real, xie2023desra} is introduced in SR task to avoid generating over-smooth results. 
BSRGAN~\cite{BSRGAN} designs a random shuffle strategy to enlarge the degradation space for training a comprehensive SR model. 
Real-ESRGAN~\cite{realesrgan} presents a more practical degradation process called ``high-order'' to synthesize realistic LR images. 
KDSRGAN~\cite{KDSR} estimates the implicit degradation representation to assist the restoration process. 
While GAN-based BSR methods require only one step to restore the LR image, their capability to super-resolve complex natural images is limited.
In this work, our~\name~seamlessly attains superior restoration performance based on diffusion model, making it a compelling choice. 

\noindent \textbf{Diffusion-based BSR.}
Diffusion models has demonstrated significant advantages in image generation tasks (\textit{e.g.}, text-to-image). 
One common approach~\cite{yue2023resshift, sr+, xia2023diffir, wang2024sinsr}x is training a non-multimodal diffusion model from scratch, which takes the concatenation of a LR image and noise as input in every step. 
Another approach~\cite{PASD, DiffBIR, wu2023seesr, StableSR, sun2023improving} fully leverages the prior knowledge from a pre-trained multimodal diffusion model (\textit{i.e.}, SD model), which requires training a ControlNet and incorporates new adaptive structures (\textit{e.g.}, cross-attention). 
SD-based methods excel in performance compared to the aforementioned approaches, as they effectively incorporate high-level information. 
However, the large number of model parameters and the need for numerous sampling steps pose substantial challenges to application in the real world.

\noindent \textbf{Efficient Diffusion Models.}
Several works~\cite{lu2022dpm, ddim,liu2022pseudo,dpm++, sdxl-lighting, sdxs} are proposed to accelerate the inference process of DM. 
Although these methods can reduce the sampling steps from thousands to $20$-$50$, the restoration effect will deteriorate dramatically. 
Recnet, adversarial diffusion distillation~\cite{ADD} is proposed to achieve $1$$\sim$$4$ steps inference while maintaining satisfactory generating ability. 
However, ADD was originally designed for the text-to-image task. 
Considering the multifaceted nature of BSR, such as image quality, degradation, or the trade-off between fidelity and realness, employing ADD to expedite the SD-based model for BSR is non-trivial. 
In contrast,~\name~introduces two pivotal designs to adapt ADD into BSR tasks, making it both effective and efficient.

\begin{figure*}[t!]
    \centering
    \includegraphics[width=1\linewidth]{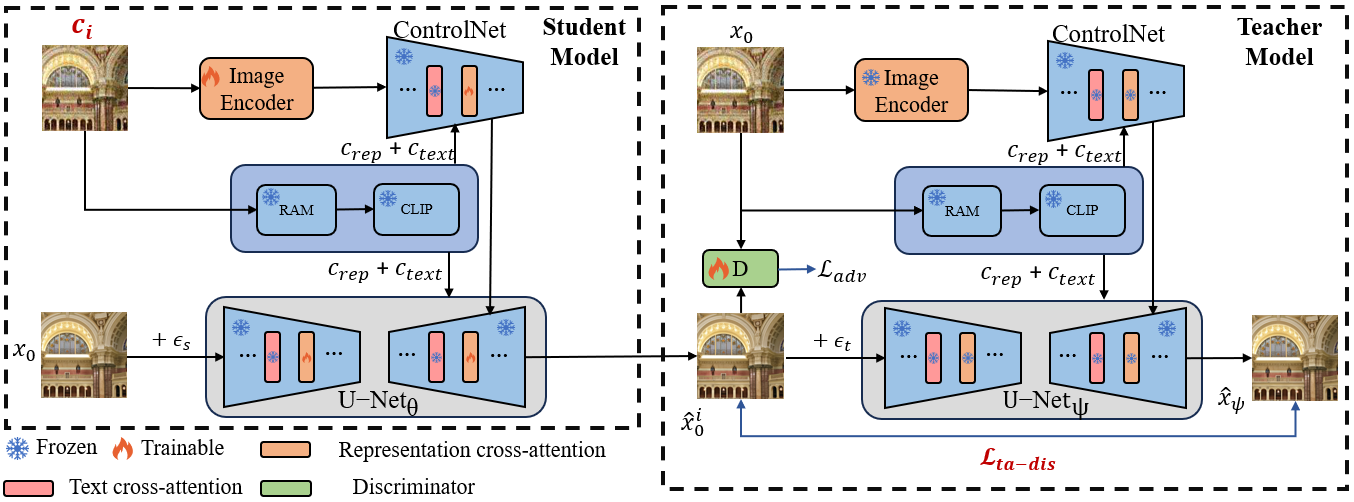}
    \caption{\textbf{Overview of~\name}. Our proposed~\name~consists of a student model, a pretrained teacher model, and a discriminator. Let $\mathcal{C} = \{ x_{LR}, \hat{x}_0^1, \hat{x}_0^2, \hat{x}_0^3 \}$, where $c_i$ denotes the $i$-th element of $\mathcal{C}$, $\hat{x}_0^i$ is the predicted HR image in the $(i+1)$ inference step. In our proposed prediction-based self-refinement strategy, we use $x_{LR}$
    as the condition only in the first inference step. For the subsequent inference steps, we use the predicted HR image $\hat{x}_0^i$ from the previous step as the condition.
    }
    \label{fig:overview}
\end{figure*}

\vspace{-1.5mm}
\section{Methodology}
\label{sec:methodology}
\vspace{-1.5mm}
\subsection{Overview of AddSR}
\label{overview}
\textbf{Network Components.} 
The AddSR training procedure primarily consists of three components: the student model with weights $\theta$, the pretrained teacher model with frozen weights $\psi$ and the discriminator with weights $\phi$, as depicted in Fig. \ref{fig:overview}. Specifically, both the student model and the teacher model share identical structures, with the student model initialized from the teacher model. The student model incorporates a ControlNet \cite{controlnet} to receive $x_{LR}$ or predicted $\hat{x}_0^{i-1}$ for controlling the output of the U-Net \cite{U-net}. Furthermore, the student model utilizes RAM \cite{RAM} to obtain representation embeddings $c_{rep}$, extracting high-level information (\textit{i.e.}, image content) and sends this information to CLIP \cite{clip} to generate text embeddings $c_{text}$. These embeddings help the backbone (U-Net and ControlNet) produce high-quality restored images. As for the discriminator, we follow StyleGAN-T \cite{styleganT} conditioned on $c_{img}$ extracted from $x_{LR}$ by DINOv2 \cite{dinov2}.

\begin{figure*}[t!]
	\centering
	\includegraphics[width=1\linewidth]{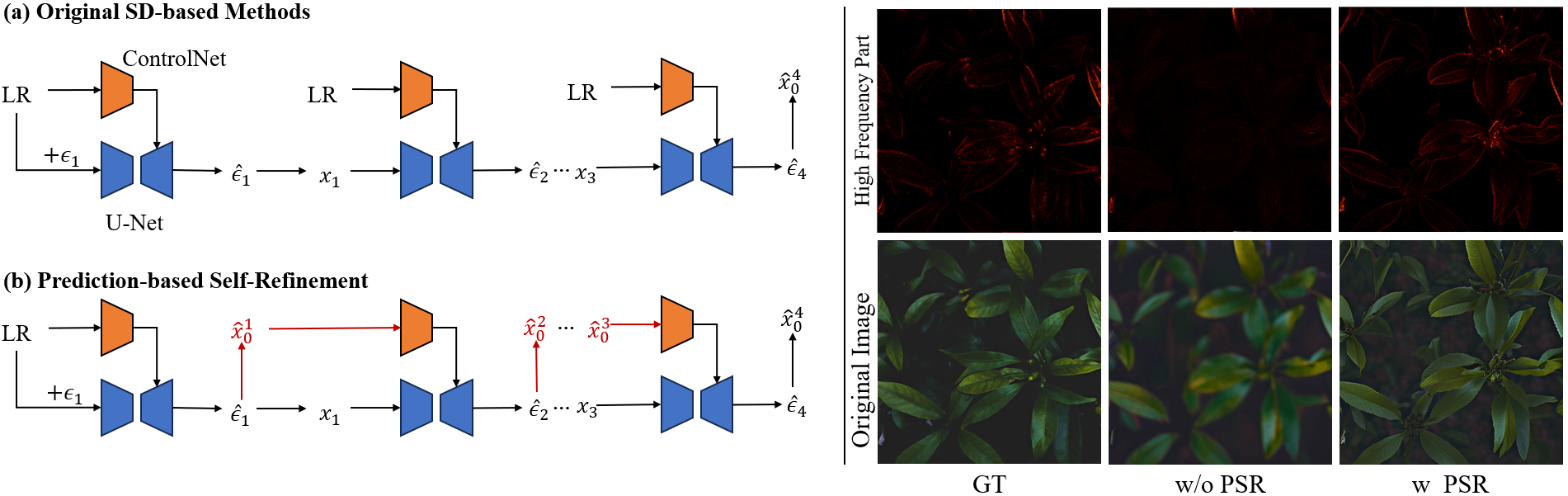}
        \vspace{-3mm}
	\caption{\textbf{Illustration of the proposed PSR}. 
		The previous SD-based methods usually use LR image to guide model's output, 
		while our PSR additionally utilizes the predicted HR image from the previous step to provide better supervision with marginal additional time cost.}
	\label{fig:PSR}
\end{figure*}

\begin{figure*}[t]
    \centering
    \includegraphics[width=1\linewidth]{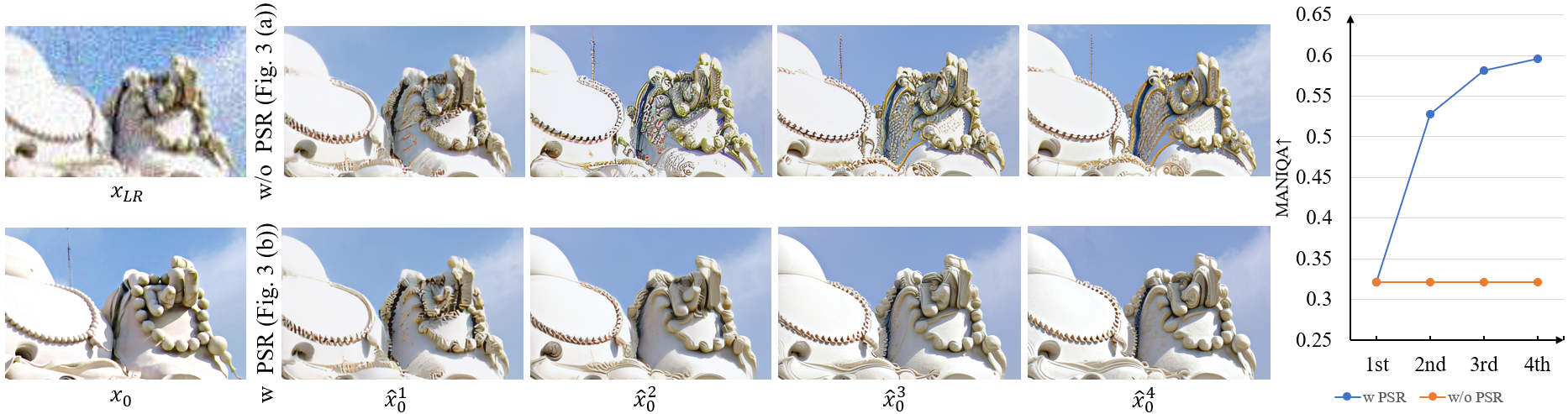}
    \vspace{-3mm}
    \caption{\textbf{Left}: Visual comparisons with and without PSR.
            \textbf{Right}: Perception quality of the control signal at each timestep. MANIQA is calculated between the input of ControlNet and $x_0$.
            }
\label{fig:PSR_illu}
\end{figure*}

\noindent \textbf{Training Procedure}.
(1) \textbf{Student model with prediction-based self-refinement}. Firstly, we uniformly choose a student timestep $s$ from $\{s_1, s_2, s_3, s_4\}$ (evenly selected from 0 to 999) and employ the forward process on the HR image $x_0$ to generate the noisy state $x_s = \sqrt{\overline{\alpha}_s}x_0 + \sqrt{1-\overline{\alpha}_s}\epsilon$. 
Secondly, we input $x_s$ with the condition $c_i$, the $i$-th element of $\mathcal{C} = \{ x_{LR}, \hat{x}_0^1, \hat{x}_0^2, \hat{x}_0^3 \}$ ($\hat{x}_0^{i-1}$ is obtained by PSR to reduce the degradation impact and provide more high-frequency information to the restoration process, as detailed in Sec. \ref{PSR}), along with $c_{rep}$ and $c_{text}$, into the student model to generate samples $\hat{x}_0^i(x_s, s, c_{rep}, c_{text}, c_i)$. 
(2) \textbf{Teacher model.} Firstly, we equally choose a teacher timestep $t$ from $\{t_1, t_2, ..., t_{1000}\}$ and employ forward process to student-generated samples $\hat{x}_\theta$ to obtain the noisy state $\hat{x}_{\theta,t} = \sqrt{\overline{\alpha}_t}\hat{x}_0^i + \sqrt{1-\overline{\alpha}_t}\epsilon$. Secondly, we put $\hat{x}_{\theta,t}$ with condition $x_0$, $c'_{rep}$ and $c'_{text}$ into teacher model to generate samples $\hat{x}_\psi(\hat{x}_{\theta,t}, t, c'_{rep}, c'_{text}, x_0)$. Note that \textit{$\hat{x}_\psi$ is conditioned on $x_0$ instead of $x_{LR}$}. The primary reason is that substituting $x_{LR}$ with $x_0$ to regulate the output of the teacher model can force student model implicitly learning the high-frequency information of HR images even conditioned on $c_i$. 
(3) \textbf{Timestep-adaptive ADD} for BSR task. It consists of two parts: adversarial loss and a novel timestep-adaptive distillation loss, which is correlated with both the teacher and student model timesteps. The overall objective is:
\begin{equation}
\begin{split}
\mathcal{L}_{TA-ADD} = &\mathcal{L}_{ta-dis}(\hat{x}_0^i(x_s, s, \rho, c_i), \hat{x}_\psi(\hat{x}_{\theta,t}, t, \rho', x_0), \\ d(s, t)) +
& \lambda\mathcal{L}_{adv}(\hat{x}_0^i(x_s, s, \rho, c_i), x_0, \psi_{c_{img}}),
\end{split}
\end{equation}
\noindent where $\rho$ denotes the $c_{rep}$ and $c_{text}$, $\rho'$ stands for $c'_{rep}$ and $ c'_{text}$. $\lambda$ is the balance weight, empirically set to 0.02. $\psi_{c_{img}}$ is the discriminator conditioned on $c_{img}$. $d(s, t)$ is a weighting function defined by student timestep $s$ and teacher timestep $t$, dynamically adjusting $\mathcal{L}_{ta-dis}$ and $\mathcal{L}_{adv}$ to alleviate perception-distortion imbalance. Further analysis is provided in Sec.~\ref{TA loss}.

\subsection{Prediction-based Self-\\Refinement} 
\label{PSR}
\textbf{Motivation.}
As shown in Fig.~\ref{fig:PSR} (a), original SD-based methods directly use LR images to control the output of DM in each inference step. 
However, some studies~\cite{DiffBIR, StableSR, liao2024denoisingadaptationnoisespacedomain} have found that the restored results can be affected by the condition quality, as LR images often suffer from multiple degradations, which can significantly disrupt the restoration process (\textit{e.g.}, see the first line of Fig.~\ref{fig:PSR_illu}). 
To provide a better condition, these methods employ \textit{additional degradation removal models} to pre-clean LR images, aiming to mitigate the impact of degradation. 
However, such approaches often compromise efficiency, which hinders designing an efficient method.

\begin{figure*}[t]
    \centering
    \includegraphics[width=1\linewidth]{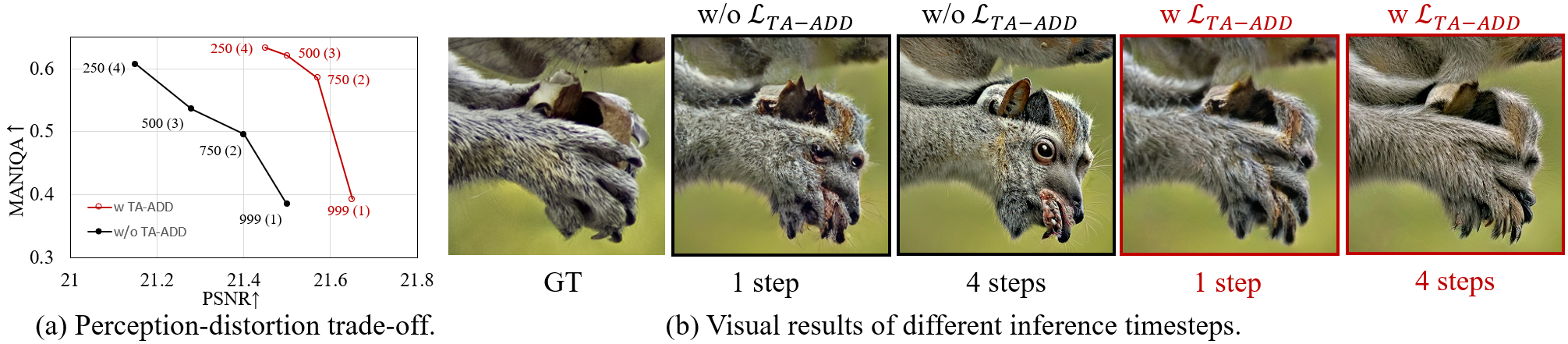}
    \vspace{-3mm}
    \caption{\textbf{Illustrations of TA-ADD in perception-distortion trade-off}. 
    (a) The perception and fidelity variation trends with and without TA-ADD. 
    MANIQA and PSNR stand for perception quality and fidelity, respectively. 
    (b) The outputs at 1st and 4th timesteps. 
    The final output without $\mathcal{L}_{TA-ADD}$ hallucinates the paw into an animal head, while~\name~retains the appearance of the paw.}
    \label{fig:ta loss1}
\end{figure*}

\begin{figure}[t]
    \centering
    \includegraphics[width=1\linewidth]{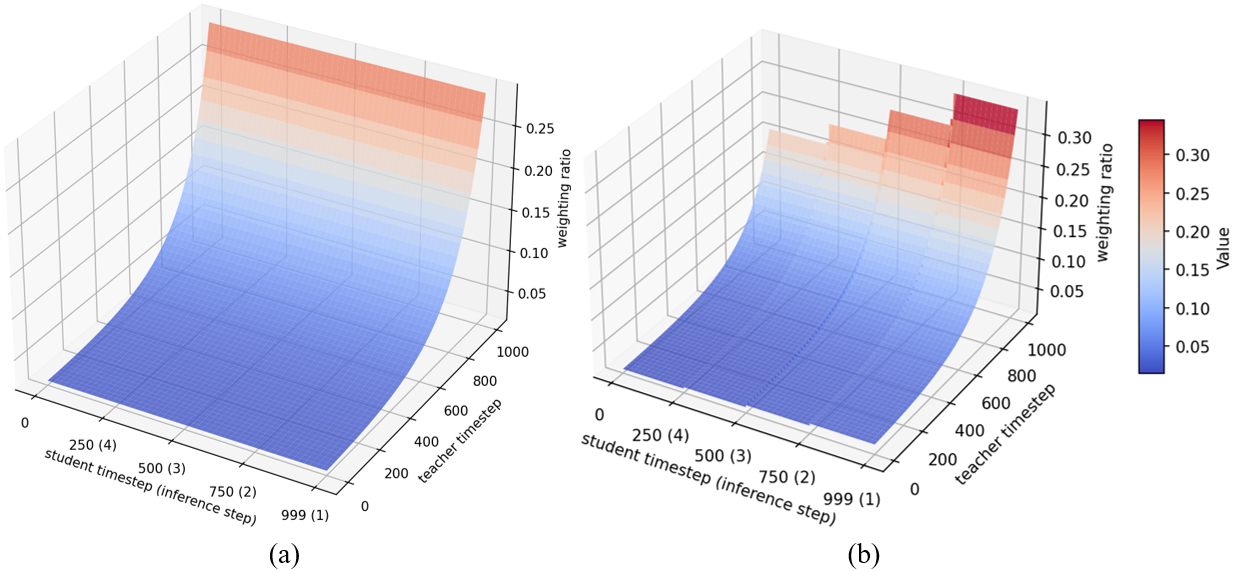}
    \vspace{-3mm}
    \caption{\textbf{Relation between weighting ratio and timesteps}. (a) weighting ratio = $\lambda/(\prod_{i=0}^{t}(1-\beta_t))^\frac{1}{2}$. Once the teacher timestep is established, the weighting ratio remains constant. (b) weighting ratio = $\lambda/d(s, t)$. 
		Even the teacher timestep is established, the weighting ratio can change to balance perception-distortion across different student timesteps.}
    \label{fig:weighting_ratio}
\end{figure}

\noindent \textbf{Approach.}
To achieve efficient restoration of high-frequency details, we propose a novel prediction-based self-refinement strategy, which incurs only minimal efficiency overhead. The idea of PSR is to utilize the predicted noise to estimate HR.
Specifically, we use the following equation: 
\begin{equation}
   \hat{x}_0 = (x_s-\sqrt{1-\overline{\alpha}_s}\epsilon_{\theta,s}) / \sqrt{\overline{\alpha}_s}
   \label{eq4}
\end{equation}
to estimate the HR image $\hat{x}_0$ from predicted noise in each step, and then control the model output in next step, where $x_s$ is the noisy state and $\epsilon_{\theta,s}$ is the predicted noise at timestep $s$. 
The $\hat{x}_0$ in each step has more high-frequency information to better control the model output (\textit{e.g.}, Fig.~\ref{fig:PSR}-right and also Fig.~\ref{fig:PSR_illu}-left).
Although PSR does not use additional modules to pre-clean LR image, the HR image estimated by PSR exhibit superior quality compared to the LR image (Fig.~\ref{fig:PSR_illu}-right).
By leveraging our simple yet effective PSR,~\name~captures conditions with high-frequency information, generating restored results with enhanced detail, without sacrificing efficiency.

\subsection{Timestep-Adaptive ADD}
\label{TA loss}
\textbf{Motivation.}
Perception-distortion trade-off~\cite{perception} is a well-known phenomenon in SR task. 
We observe that training BSR task with ADD directly exacerbates this phenomenon, as shown in Fig.~\ref{fig:ta loss1}(a).
Specifically, during the first three inference steps, there is a significant decrease in fidelity, accompanied by improvement in perception quality. 
In the last inference step, fidelity remains at a low level, while perception quality undergoes a dramatic increase. 
The aforementioned scenario may give rise to two issues: \textit{(1) When the inference step is small, the quality of restored image is subpar. (2) As the inference step increases, the generated images may exhibit ``hallucinations''.}

\noindent \textbf{Analysis.}
The primary reason lies with ADD, which maintains a consistent weight for GAN loss and distillation loss across various student timesteps, as depicted in Fig.~\ref{fig:weighting_ratio}(a).
 Once the teacher timestep is established, the ratio of adversarial loss and distillation loss remains constant for different student timesteps.
However, since the perception quality of generated images gradually increases with larger inference steps, the weight-invariant ADD may lead to insufficient adversarial constraints on the student model during small inference steps, resulting in the generation of blurry images.
Conversely, as the inference step increases, the adversarial training constraints become too strong, leading to the generation of ``hallucinations" (see Fig.~\ref{fig:ta loss1}(b)). 

\begin{table*}[t!]
	\centering
	\caption{Quantitative comparison with SotAs on different degradation cases. 
		`*' indicates that the metric is non-reference. 
		The best results are marked in \textcolor{red}{red}, while the second best ones are in \textcolor{blue}{blue}.}
	\label{tab:systhetic data}
        \resizebox{\textwidth}{!}{
        \begin{tabular}{c c|ccccc|ccccccc}
			\hline
			\multirow{2}{*}{Datasets} & \multirow{2}{*}{Metrics} & BSRGAN & Real-ESRGAN & MM-RealSR & LDL & FeMaSR & StableSR-200 & ResShift-15 & PASD-20 & DiffBIR-50 & SeeSR-50 & \name-$1$ & \name-$4$\\ 
			~ & ~ & ICCV 2021 & ICCVW 2021 & ECCV 2022 & CVPR 2022 & MM 2022 & IJCV 2024 & NeurIPS 2023 & ECCV 2024 & ECCV 2024 & CVPR 2024 & - & - \\ \hline
			\multirow{6}{*}{\makecell[l]{Single: \\SR($\times$4)}} & MANIQA$^*\uparrow$ & 0.3990 & 0.3859 & 0.3959 & 0.3501 & 0.4603 & 0.4088 & 0.4582 & 0.4405 & 0.4680 & \textcolor{blue}{0.5082} & 0.3894 & \textcolor{red}{0.6430} \\
			~ & MUSIQ$^*\uparrow$ & 66.06 & 63.32 & 64.22 & 61.10 & 65.31 & 65.46 & 65.50 & 66.80 & 67.61 & \textcolor{blue}{68.88} & 63.05 & \textcolor{red}{71.43} \\
			~ & CLIPIQA$^*\uparrow$ & 0.5951 & 0.5367 & 0.5967 & 0.5120 & 0.6773 & 0.6483 & 0.6803 & 0.6396 & 0.6934 & \textcolor{blue}{0.7039} & 0.5572 & \textcolor{red}{0.7794} \\
			~ & NIQE$^*\downarrow$& 5.01 & 5.21 & 5.22 & 5.39 & 5.80 & 5.35 & 5.74 & \textcolor{red}{4.68} & 4.88 & 5.06 & 5.31 & \textcolor{blue}{4.75}\\
			~ & LPIPS$\downarrow$ & 0.2003 & 0.1962 & 0.1934 & 0.1892 & \textcolor{blue}{0.1770} & 0.1944 & \textcolor{red}{0.1544} & 0.1891 & 0.2388 & 0.3085 & 0.2872 & 0.2812 \\
			~ & PSNR$\uparrow$ & \textcolor{blue}{25.52} & 25.30 & 24.35 & 25.09 & 23.74 & 24.45 & \textcolor{red}{25.53} & 25.15 & 23.43 & 24.61 & 22.70 & 21.83 \\
			~ & SSIM$\uparrow$ & 0.7091 & 0.7158 & \textcolor{blue}{0.7232} & \textcolor{red}{0.7282} & 0.6788 & 0.6904 & 0.7206 & 0.6896 & 0.6025 & 0.6709 & 0.6012 & 0.5651 \\ \hline
			\multirow{6}{*}{\makecell[l]{Mixture: \\Blur($\sigma$=2)+\\SR($\times$4)}} & MANIQA$^*\uparrow$ & 0.3823 & 0.3688 & 0.3796 & 0.3337 & 0.4184 & 0.3587 & 0.4195 & 0.4124 & 0.4648 & \textcolor{blue}{0.4974} & 0.3779 & \textcolor{red}{0.6340} \\
			~ & MUSIQ$^*\uparrow$ & 64.73 & 60.89 & 62.21 & 58.64 & 62.96 & 60.85 & 62.02 & 64.41 & 67.09 & \textcolor{blue}{68.27} & 61.95 & \textcolor{red}{71.11} \\
			~ & CLIPIQA$^*\uparrow$ & 0.5752 & 0.5116 & 0.5687 & 0.4910 & 0.6390 & 0.5819 & 0.6375 & 0.6026 & 0.6857 & \textcolor{blue}{0.6892} & 0.5389 & \textcolor{red}{0.7727} \\
			~ & NIQE$^*\downarrow$& \textcolor{blue}{5.17} & 5.54 & 5.66 & 5.72 & 5.62 & 5.96 & 6.25 & \textcolor{red}{5.01} & 5.18 & 5.32 & 5.81 & 6.11\\
			~ & LPIPS$\downarrow$ & 0.2240 & 0.2267 & 0.2295 & 0.2226 & \textcolor{red}{0.1979} & 0.2384 & \textcolor{blue}{0.2029} & 0.2223 & 0.2522 & 0.2124 & 0.3007 & 0.2953 \\
			~ & PSNR$\uparrow$ & \textcolor{red}{25.07} & 24.74 & 24.20 & 24.45 & 24.00 & 24.01 & \textcolor{blue}{24.95} & 24.70 & 22.97 & 24.12 & 22.57 & 21.69 \\
			~ & SSIM$\uparrow$ & 0.6820 & 0.6890 & \textcolor{blue}{0.6927} & \textcolor{red}{0.6973} & 0.6730 & 0.6596 & 0.6926 & 0.6688 & 0.5802 & 0.6508& 0.5905 & 0.5556 \\ \hline
			\multirow{6}{*}{\makecell[l]{Mixture: \\SR($\times$4)+\\Noise($\sigma=$40)}} & MANIQA$^*\uparrow$ & 0.2645 & 0.3120 & 0.3285 & 0.3138 & 0.3123 & 0.3485 & 0.3741 & 0.4270 & 0.4121 & \textcolor{blue}{0.5537} & 0.4320 & \textcolor{red}{0.6517} \\
			~ & MUSIQ$^*\uparrow$ & 50.47 & 53.43 & 56.53 & 53.30 & 56.55 & 52.24 & 60.99 & 64.20 & 61.85 & \textcolor{blue}{70.32} & 65.54 & \textcolor{red}{71.26} \\
			~ & CLIPIQA$^*\uparrow$ & 04543 & 0.4761 & 0.5158 & 0.6208 & 0.5178 & 0.4414 & 0.5949 & 0.5503 & 0.6149 & \textcolor{blue}{0.7557} & 0.6219 & \textcolor{red}{0.7768} \\
			~ & NIQE$^*\downarrow$& 7.04 & 6.00 & \textcolor{blue}{4.40} & 5.61 & \textcolor{red}{4.27} & 5.12 & 6.10 & 5.02 & 5.09 & 4.95 & 4.87 & 6.29 \\
			~ & LPIPS$\downarrow$ & 0.4611 & 0.3601 & \textcolor{blue}{0.3052} & 0.3138 & 0.3267 & 0.4017 & 0.3129 & 0.3451 & 0.3404 & \textcolor{red}{0.2999} & 0.3546 & 0.3488 \\
			~ & PSNR$\uparrow$ & 17.90 & 21.97 & 22.04 & \textcolor{blue}{22.68} & 21.84 & 21.20 & \textcolor{red}{22.78} & 22.12 & 22.22 & 21.04 & 21.01 & 20.79 \\
			~ & SSIM$\uparrow$ & 0.5210 & \textcolor{red}{0.6044} & \textcolor{blue}{0.5998} & 0.5838 & 0.5421 & 0.5077 & 0.5979 & 0.5587 & 0.5311 & 0.5388 & 0.5684 & 0.5621 \\ \hline
			\multirow{6}{*}{\makecell[l]{Mixture: \\Blur($\sigma=$2)+\\SR($\times$4)+\\Noise($\sigma=$20)+\\JPEG(q=50)}} & MANIQA$^*\uparrow$ & 0.3524 & 0.3374 & 0.3287&0.3082	&0.3271	&0.3452	&0.3702	&0.4024	&0.4538	&\textcolor{blue}{0.5266}	& 0.3930 &\textcolor{red}{0.6335}\\
			~ & MUSIQ$^*\uparrow$ & 59.83 &55.54 &55.30&	52.79 &60.87&61.21&	56.99&63.25&64.50&\textcolor{blue}{69.08}&62.69 &\textcolor{red}{70.59} \\
			~ & CLIPIQA$^*\uparrow$ & 0.5380 &0.5047 &	0.4978 	&0.4699 &0.6061 &	0.6010 &0.5888&0.5733&0.6626&\textcolor{blue}{0.7180} &	0.5669 & \textcolor{red}{0.7703} \\
			~ & NIQE$^*\downarrow$& 5.31 & 5.69 & 5.70 & 5.77 & \textcolor{blue}{4.87} & 6.33 & 7.03 & 5.49 & 4.93 & 5.06 & 5.13 & \textcolor{red}{4.68}\\
			~ & LPIPS$\downarrow$ & 0.3223 & 0.3346 & 0.3372 & 0.3272 & \textcolor{red}{0.2922} & 0.3429 & 0.3526 & 0.3482 & 0.3502 & \textcolor{blue}{0.3085} & 0.3398 & 0.3368 \\
			~ & PSNR$\uparrow$ &\textcolor{red}{23.04}&\textcolor{blue}{22.70}&22.47&22.36&22.17&22.39&22.36&22.25&21.46&21.86&21.65&21.45 
			\\
			~ & SSIM$\uparrow$ & 0.5866&0.5935&\textcolor{red}{0.5950}&\textcolor{blue}{0.5948}&0.5633&0.5704&0.5574&0.5594&0.5029&0.5474&0.5312&0.5210 
			\\ \hline
	\end{tabular}}
\end{table*}

\noindent \textbf{Approach.}
To address this issue, we extend the original unary weighting function $(\prod_{i=0}^{t}(1-\beta_t))^\frac{1}{2}$ to a bivariate weighting function $d(s,t)$, allowing for dynamic adjustment of the ratio between adversarial loss and distillation loss based on both student timestep and teacher timestep, as shown in Fig.~\ref{fig:weighting_ratio}(b). 
Specifically, we increase this ratio when only one inference step is performed, and gradually decrease it as the inference step increases. 
This alleviates the aforementioned issue of generating blurry images with small inference step and ``hallucinations" with larger inference steps. We employ the exponential forms to control the weighting ratio. The function $d(s,t)$ is:
\begin{equation}
    d(s, t) = (\prod_{i=0}^{t}(1-\beta_t))^\frac{1}{2} \times \mu \cdot \nu^{p(s) - 1},
    \label{eq:exponential}
\end{equation}
where $\beta$ represents the noise schedule coefficient, with $t$ and $s$ denoting the teacher timestep and student timestep, respectively. 
The hyper-parameter $\mu$ sets the initial weighting ratio, 
while $\nu$ controls the distillation loss increase over student timesteps, typically resulting in higher fidelity with larger $\nu$.
The function $p(\cdot)$ serves as a projection function that maps student timesteps to inference steps (e.g., mapping $s=999$ to $1$). We primarily consider the exponential and linear forms to control the weighting ratio. A comparison of detailed settings for different hyper-parameters are provided in Sec. \ref{ablation study}. From these comparisons, we find that the exponential form of $d(s,t)$ yields good results, so we use Eq.~(\ref{eq:exponential}) as the distillation loss function for the remaining experiments.

\section{Experiments}
\label{sec:experiments}
\subsection{Experimental Settings}
\textbf{Training Datasets.}
We adopt DIV2K~\cite{DIV2K}, Flickr2K~\cite{Flickr2K}, first $20$K images from LSDIR~\cite{lsdir} and first $10$K face images from FFHQ~\cite{ffhq} for training. We use the same degradation model as Real-ESRGAN~\cite{realesrgan} to synthesize HR-LR pairs. 

\noindent \textbf{Test Datasets.}
We evaluate~\name~on $4$ datasets: DIV2K-val~\cite{DIV2K}, DRealSR~\cite{DrealSR}, RealSR~\cite{realsr} and RealLR200~\cite{wu2023seesr}. 
We conduct $4$ degradation types on DIV2K-val to comprehensively assess~\name, and except RealLR200, all datasets are cropped to $512$$\times$$512$ and degraded to $128$$\times$$128$ LR image. 

\noindent \textbf{Implementation Details.}
We adopt SeeSR~\cite{wu2023seesr} as the teacher model. 
Note that our approach is applicable to most of the existing SD-based BSR methods for improving restoration results and acceleration. 
The student model is initialized from the teacher model, and fine-tuned with Adam optimizer for $50$K iterations. 
The batch size and learning rate are set to $6$ and $2$$\times10^{-5}$.~\name~is trained under $512$$\times$$512$ resolution with $4$ NVIDIA A$100$ GPUs ($40$G).

\begin{figure}[t!]
	\centering
	\includegraphics[width=1\linewidth]{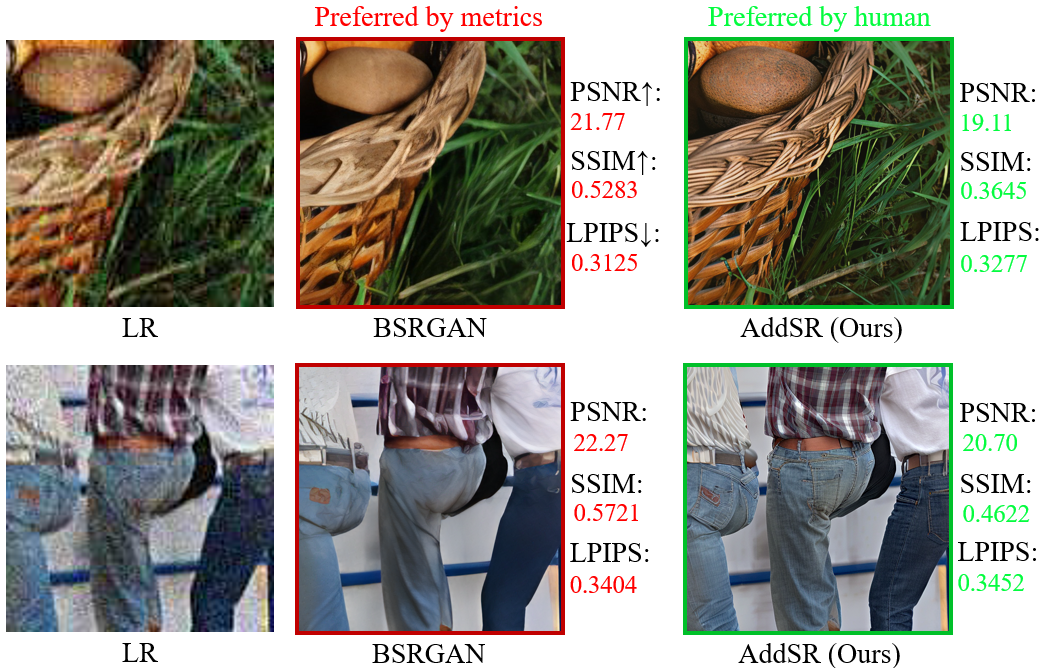}\
	\caption{\textbf{Illustration on disparity} between full-reference metrics and human preference. Despite AddSR achieves lower scores in full-reference metrics, it generates human-preferred images.}
	\label{fig:explanation}
\end{figure}

\begin{figure*}[t!]
    \centering
    \includegraphics[width=1\linewidth]{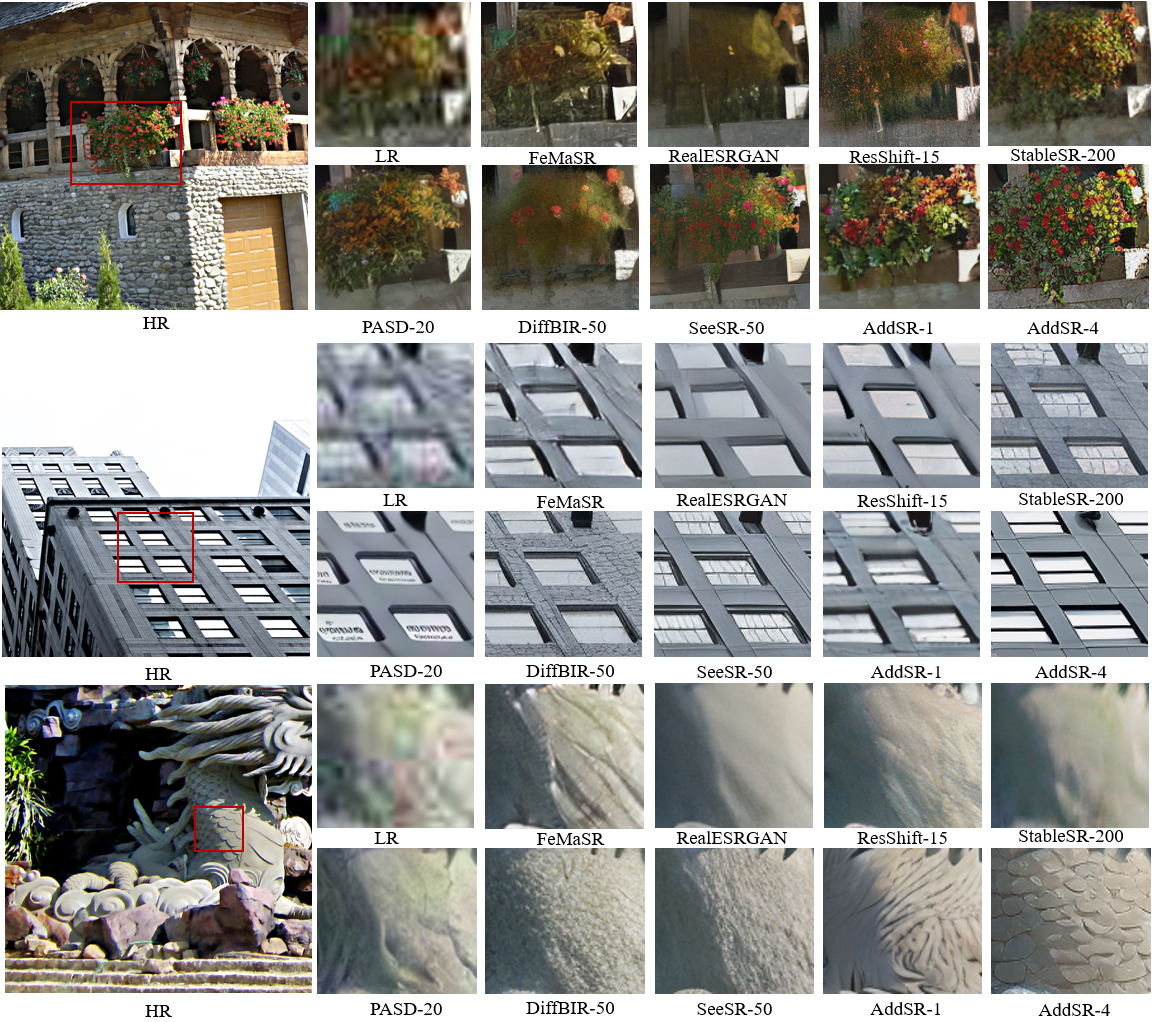}
    \vspace{-3mm}
    \caption{\textbf{Visual comparisons on synthetic LR images}. Please zoom in for a better view.}
    \label{fig:synthetic}
\end{figure*} 

\begin{table*}[t!]
    \caption{Quantitative comparison of SUPIR: Inference time, model size, training source, and metrics.}
    \label{tab:comparison with supir}
    \resizebox{\textwidth}{!}{
    \begin{tabular}{ccccccccc}
       \hline
       Model & \#Params [B] & Time [s] & Training Source & Dataset Size [M] & PSNR$\uparrow$ & SSIM$\uparrow$ & MANIQA$\uparrow$ & CLIPIQA$\uparrow$\\ \hline
        SUPIR (CVPR 2024) & $\sim$15.56 & 14.17 & 64 A6000 (48G) & 20 & 20.78 & 0.4587 & \bf{0.6787} & \bf{0.7992} \\
        AddSR (Ours) & $\sim$\bf{2.28} & \bf{0.80} & \bf{4 A100 (40G)} & \bf{0.034} & \bf{21.45} & \bf{0.5210} & 0.6335 & 0.7703 \\ \hline
    \end{tabular}}
\end{table*}

\begin{figure}[t!]
    \centering
    \includegraphics[width=1\linewidth]{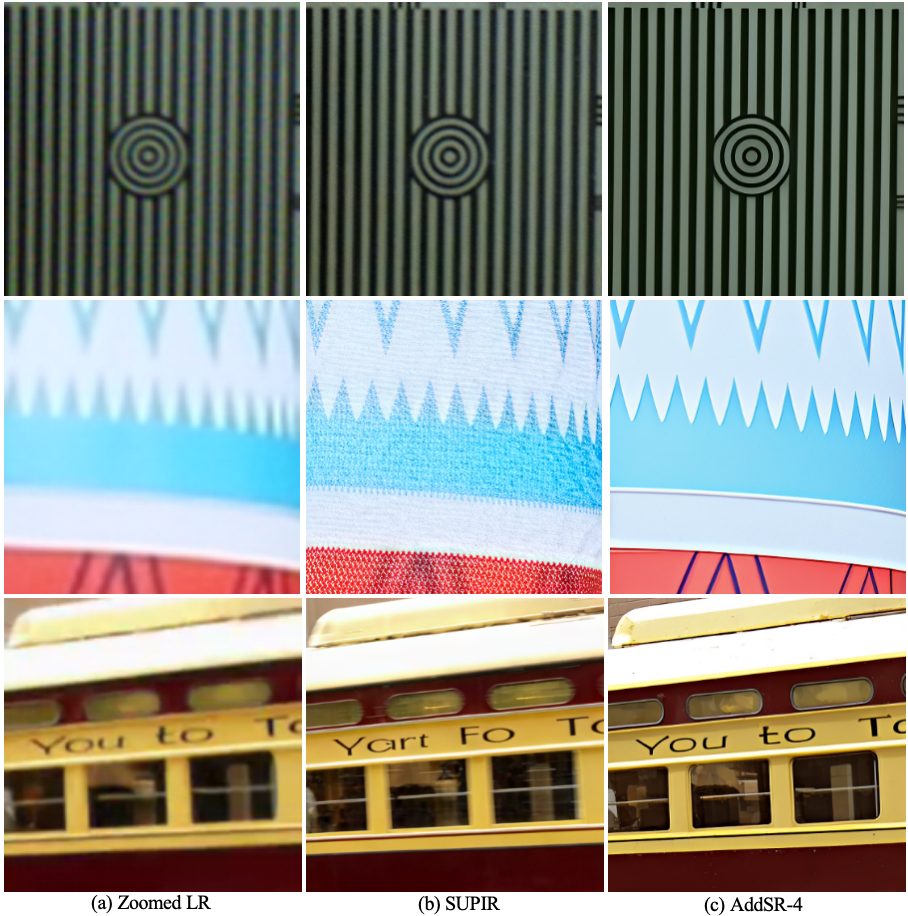}
    \vspace{-3mm}
    \caption{\textbf{Visual comparisons with SUPIR}. Please zoom in for a better view.}
    \label{fig:supir_comparison}
\end{figure}

\noindent \textbf{Evaluation Metrics.}
We employ non-reference metrics (\textit{i.e.}, MANIQA~\cite{yang2022maniqa}, MUSIQ~\cite{musiq}, CLIPIQA~\cite{clipiqa}) and reference metrics (\textit{i.e.}, LPIPS~\cite{LPIPS}, PSNR, SSIM~\cite{ssim}) to comprehensively evaluate~\name. 
Non-reference metrics are prioritized as they closely align with human perception.

\noindent \textbf{Compared Methods.}
Extensive state-of-the-art BSR methods are compared, including GAN-based methods: BSRGAN~\cite{BSRGAN}, Real-ESRGAN~\cite{realesrgan}, MM-RealSR~\cite{mou2022mmrealsr}, LDL~\cite{LDL}, FeMaSR~\cite{femasr} and diffusion-based methods: StableSR~\cite{StableSR}, ResShift~\cite{yue2023resshift}, PASD~\cite{PASD}, DiffBIR~\cite{DiffBIR}, SeeSR~\cite{wu2023seesr}.

\subsection{Evaluation on Synthetic Data}
\label{Sec: Synthetic Data}
To demonstrate the superiority of the proposed~\name~in handling various degradation cases, we synthesized $4$ test datasets using the DIV2K-val dataset with different degradation processes. The quantitative results are summarized in Tab.~\ref{tab:systhetic data}. Since SD-based methods emphasize perceptual quality, we provide results using perceptual-priority parameters. In the ablation study (Tab. \ref{tab:fidelity}), we present the corresponding results under balanced parameters.
The conclusions include:
($1$) Our~\name-$4$ achieves the highest scores in MANIQA, MUSIQ and CLIPIQA across $4$ degradation cases.
Especially for MANIQA,~\name~surpasses the second-best method by more than 16$\%$ on average. 
($2$) 
Diffusion-based models usually achieve low scores in full-reference metrics like PSNR, SSIM and LPIPS, possibly because of their powful generative ability for realistic details that do not exist in GT.
However, full-reference metrics cannot precisely reflect human preferences (see Fig.~\ref{fig:explanation}), as discussed in~\cite{SUPIR, pipal, gu2022ntire}.
($3$)~\name-$1$ can generate comparative results against other SD-based methods except SeeSR, but significantly reduces the sampling steps (\textit{i.e.}, from $\geq$$15$ steps to only $1$ step).

For a more intuitive comparison, we provide visual results in Fig.~\ref{fig:synthetic}. 
One can see that GAN-based method like FeMaSR fails to reconstruct the clean and detailed HR images of the three displayed LR images.
As for SD-based method DiffBIR, it tends to generate wrong texture. 
This is mainly because DiffBIR uses a degradation removal structure to remove the degradation of LR images. 
However, the processed LR image is blurry, which may lead to the blurry results. 
Thanks to our proposed PSR,~\name~uses the predicted $\hat{x}_0^{i-1}$ to control the model output, which has more high-frequency information and nearly no extra time cost. 
With TA-ADD,~\name~can generate precise images and rich details.
In a nutshell,~\name~produces images with better perceptual quality than the state-of-the-art models while requiring fewer inference steps and less time.

Moreover, we provide the comparison with SOTA perceptual method SUPIR~\cite{SUPIR} in Tab.~\ref{tab:comparison with supir}, which details parameters, inference time, training sources, training data, and metrics. As shown, SUPIR exhibits better perceptual quality compared to AddSR. However, AddSR strikes a better balance among model size, inference time, fidelity, perceptual quality and training resource consumption. Although SUPIR demonstrates powerful restoration capabilities, it may still fail in certain degradation scenarios, as shown in the first example of Fig.~\ref{fig:supir_comparison}. Furthermore, due to its strong generative ability, the restored results may include details that do not align with the original LR images, such as the texture in the second example and the English letters in the third example in Fig.~\ref{fig:supir_comparison}.

\begin{table*}[t!]
    \centering
    \caption{Quantitative comparison with state of the arts on real-world LR images.}
    \label{tab:real-world lr}
    \resizebox{\textwidth}{!}{
    \begin{tabular}{c c|ccccc|ccccccc}
    \hline
       \multirow{2}{*}{Datasets} & \multirow{2}{*}{Metrics} & BSRGAN & Real-ESRGAN & MM-RealSR & LDL & FeMaSR & StableSR-200 & ResShift-15 & PASD-20 & DiffBIR-50 & SeeSR-50 & \name-$1$ & \name-$4$\\ 
       ~ & ~ & ICCV 2021 & ICCVW 2021 & ECCV 2022 & CVPR 2022 & MM 2022 & IJCV 2024 & NeurIPS 2023 & ECCV 2024 & ECCV 2024 & CVPR 2024 & - & - \\ \hline
       \multirow{6}{*}{RealSR} & MANIQA$^*\uparrow$ & 0.3762&0.3727&0.3966&0.3417&0.3609&0.3656&0.3750&0.4041&0.4392&\textcolor{blue}{0.5396}&0.4189&\textcolor{red}{0.6597}
 \\
       ~ & MUSIQ$^*\uparrow$ & 63.28&60.36&62.94&58.04&59.06&61.11&56.06&62.92&64.04&\textcolor{blue}{69.82}&63.56&\textcolor{red}{72.25}
 \\
       ~ & CLIPIQA$^*\uparrow$ & 0.5116&0.4492&0.5281&0.4295&0.5408&0.5277&0.5421&0.5187&0.6491&\textcolor{blue}{0.6700}&0.4929&\textcolor{red}{0.7215}
 \\
       ~ & LPIPS$\downarrow$ & \textcolor{red}{0.2670} & \textcolor{blue}{0.2727} & 0.2783 & 0.2766 & 0.2942 & 0.3018 & 0.3460 & 0.3435 & 0.3658 & 0.3007 & 0.3095 & 0.3742 \\
       ~ & PSNR$\uparrow$ & \textcolor{blue}{26.49}&25.78&23.69&25.09&25.17&25.63&26.34&\textcolor{red}{26.67}&25.06&25.24&24.22&22.73 \\
       ~ & SSIM$\uparrow$ &\textcolor{red}{0.7667}&0.7621&0.7470&\textcolor{blue}{0.7642}&0.7359&0.7483&0.7352&0.7577&0.6664&0.7204&0.6863&0.6336 
 \\ \hline
       \multirow{6}{*}{DrealSR} & MANIQA$^*\uparrow$ &0.3431&0.3428&0.3625&0.3237&0.3178&0.3222&0.3284&0.3874&0.4646&\textcolor{blue}{0.5125}&0.3873&\textcolor{red}{0.6034}
 \\
       ~ & MUSIQ$^*\uparrow$ & 57.17&54.27&56.71&52.38&53.70&52.28&50.14&55.33&60.40&\textcolor{blue}{65.08}&57.42&\textcolor{red}{68.16}
 \\
       ~ & CLIPIQA$^*\uparrow$ &0.5094&0.4514&0.5171&0.4410&0.5639&0.5101&0.5287&0.5384&0.6397&\textcolor{blue}{0.6910}&0.5543&\textcolor{red}{0.7381}
 \\
       ~ & LPIPS$\downarrow$ & 0.2883 & 0.2847 & \textcolor{blue}{0.2832} & \textcolor{red}{0.2815} & 0.3169 & 0.3315 & 0.4006 & 0.3931 & 0.4599 & 0.3299 & 0.3025 & 0.3866 \\
       ~ & PSNR$\uparrow$ & 28.68&28.57&26.84&27.41&26.83&\textcolor{red}{29.14}&28.27&\textcolor{blue}{29.06}&26.56&28.09&27.49&26.09 
 \\
       ~ & SSIM$\uparrow$ & 0.8022&\textcolor{blue}{0.8042}&0.7959&\textcolor{red}{0.8069}&0.7545&0.8040&0.7542&0.7906&0.6436&0.7664&0.7588&0.7036 
 \\ \hline
       \multirow{3}{*}{RealLR200} & MANIQA$^*\uparrow$ & 0.3688&0.3656&0.3879&0.3266&0.4099&0.3672&0.4182&0.4193&0.4626&\textcolor{blue}{0.4911}&0.4215&\textcolor{red}{0.6182}
 \\
       ~ & MUSIQ$^*\uparrow$ &64.87&62.93&65.24&60.95&64.24&62.89&60.25&66.35&66.84&\textcolor{blue}{68.63}&65.02&\textcolor{red}{72.62}
 \\
       ~ & CLIPIQA$^*\uparrow$ & 0.5699&0.5423&0.6010&0.5088&0.6547&0.5916&0.6468&0.6203&\textcolor{blue}{0.6965}&0.6617&0.5679&\textcolor{red}{0.7724}
 \\ 
 \hline
    \end{tabular}}
\end{table*}

\begin{figure*}[t!]
    \centering
    \includegraphics[width=0.95\linewidth]{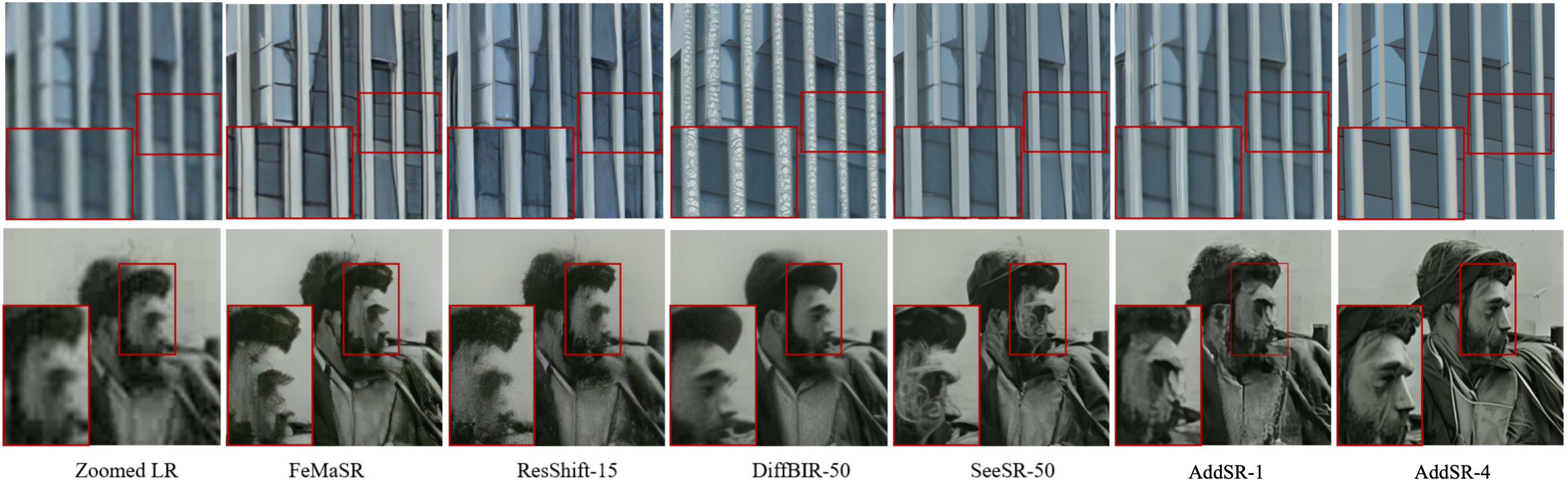}
    \caption{\textbf{Visual comparisons on real-world LR images from RealLR200}. Please zoom in for a better view.}
    \label{fig:real-world LR}
\end{figure*}

\begin{figure*}[t!]
    \centering
    \includegraphics[width=0.95\linewidth]{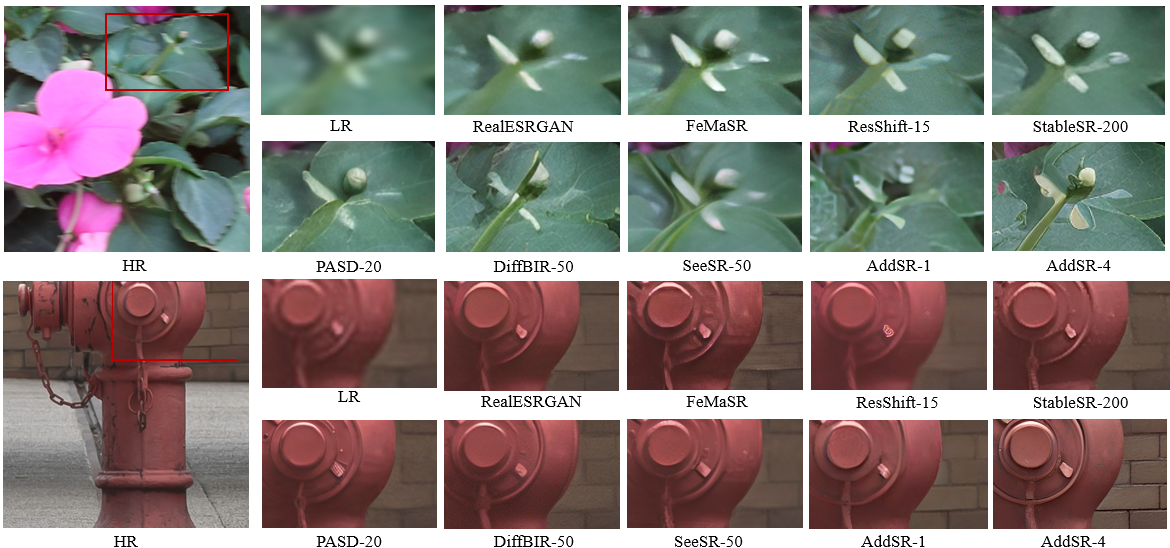}
    \caption{\textbf{Visual comparisons on real-world LR images from RealSR and DrealSR}. Zoom in for a better view.}
    \label{fig:real-world LR2}
\end{figure*}

\subsection{Evaluation on Real-World Data}
Tab.~\ref{tab:real-world lr} shows the quantitative results on $3$ real-world datasets. 
We can see that our~\name~achieves the best scores in MANIQA, MUSIQ and CLIPIQA, the same as in the synthetic degradation cases. 
This demonstrates that~\name~has an excellent generalization ability to handle unknown complex degradations, making it practical in real-world scenarios. 
Additionally,~\name-$1$ surpasses the GAN-based methods, primarily due to the integration of diffusion model with adversarial training.
This integration enables~\name~to leverage the powerful priors of the pre-trained diffusion model and inject high-level information, enhancing the restoration process and producing high-quality perceptual images, even in a \textit{one-step} inference.

\begin{figure*}[!]
	\centering
	\includegraphics[width=1\linewidth]{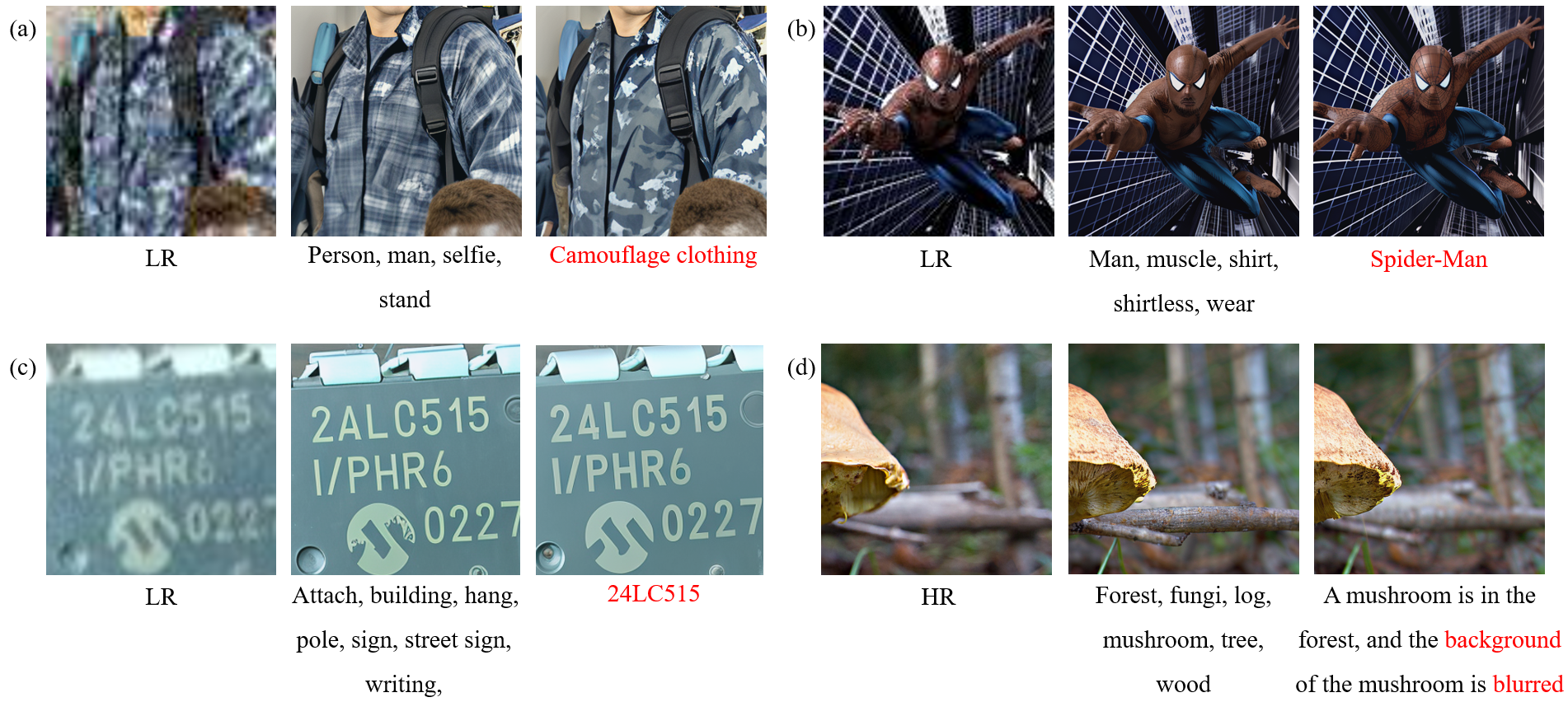}
        \vspace{-4mm}
	\caption{\textbf{Illustrations of prompt-guided restoration} that engages with manual prompts for more precise outcomes. 
		In each group, the prompts for the second and third images are obtained through RAM and manual input, respectively.
	}
	\label{fig:text_edit}
\end{figure*}

Fig.~\ref{fig:real-world LR} and Fig.~\ref{fig:real-world LR2} show the visualization results. 
We present the examples of building and face to comprehensively compare various methods. 
A noticeable observation is that~\name~generate more clear and regular line, as evidenced by the linear pattern of the building in the first example. 
In the second example, the original LR image is heavily degraded, FeMaSR and ResShift fail to generate the human face, showing only the blurry outline of the face. 
DiffBIR can generate more details, yet still unclear. 
The image generated by SeeSR exhibits artifacts. 
Conversely, our~\name~can generate comparative results with FeMaSR and ResShift in one-step.
As evaluating the inference steps,~\name~generates more clear and detailed human face, which significantly surpasses the aforementioned methods.

\vspace{2mm}
\noindent \textbf{Prompt-Guided Restoration.}
One of the advantages of diffusion model is to integrate with text. 
In Fig.~\ref{fig:text_edit}, we demonstrate that our~\name~can efficiently achieve more precise restoration results \textit{in 4 steps} by incorporating with manual prompt, \textit{i.e.}, we can manually input the text description of the LR image to assist the restoration process. 
Specifically, Fig.~\ref{fig:text_edit}(a) demonstrates that the plaid shirt in the restored image using the RAM prompt can be edited to camouflage with a manual prompt. In Fig.~\ref{fig:text_edit}(b), the Spider-Man restored with the original RAM prompt features a mouth and beard, whereas the manual prompt accurately restores the man wearing a spiderweb suit. Fig.~\ref{fig:text_edit}(c) shows that the text on the chip can be corrected from ‘2ALC515’ to ‘24LC515’ using the manual prompt. Finally, Fig.~\ref{fig:text_edit}(d) illustrates that while the RAM prompt renders the tree branches sharply, the manual prompt preserves the intended background blur of the mushroom, aligning with the Ground Truth.

\begin{table*}[t!]
    \caption{Quantitative comparison with other efficient methods, \textit{i.e.,} SeeSR-Turbo \cite{wu2023seesr}, StableSR-Turbo \cite{StableSR}, S3Diff \cite{zhang2024degradation} and SinSR \cite{wang2024sinsr}. The number following the methods denotes the number of inference steps.}
    \centering
    \label{tab:turbo}
    \resizebox{0.7\textwidth}{!}{
    \begin{tabular}{c|cc|ccc}
    \hline
        \multirow{2}{*}{Methods} & \multicolumn{2}{c}{RealLR200} & \multicolumn{3}{c}{DrealSR}\\
        ~ & MANIQA$\uparrow$ & MUSIQ$\uparrow$ & PSNR$\uparrow$ & SSIM$\uparrow$ & MANIQA$\uparrow$ \\ \hline
        SeeSR-Turbo-2 & 0.3503 & 53.88 & 26.24 & \underline{0.7531} & 0.3825 \\
        StableSR-Turbo-4 & 0.4208 & 67.55 & 24.70 & 0.6639 & 0.3381 \\
        S3Diff-1 & 0.4621 & 68.92 & 26.89 & 0.7469 & 0.4508 \\
        SinSR-1 & 0.4436 & 63.76 & \textbf{28.37} & 0.7486 & 0.3829 \\
        \hline
        \name-1 (Ours) & 0.4215 & 65.02 & 27.49 & \textbf{0.7588} & 0.3873 \\
        \name-2 (Ours) & \underline{0.5797} & \underline{71.44} & 26.78 & 0.7232 & \underline{0.5643} \\
        \name-4 (Ours) & \textbf{0.6182} & \textbf{72.62} & 26.09 & 0.7036 & \textbf{0.6034} \\
        \hline
    \end{tabular}}
\end{table*}

\begin{figure}[t!]
    \centering
    \includegraphics[width=1\linewidth]{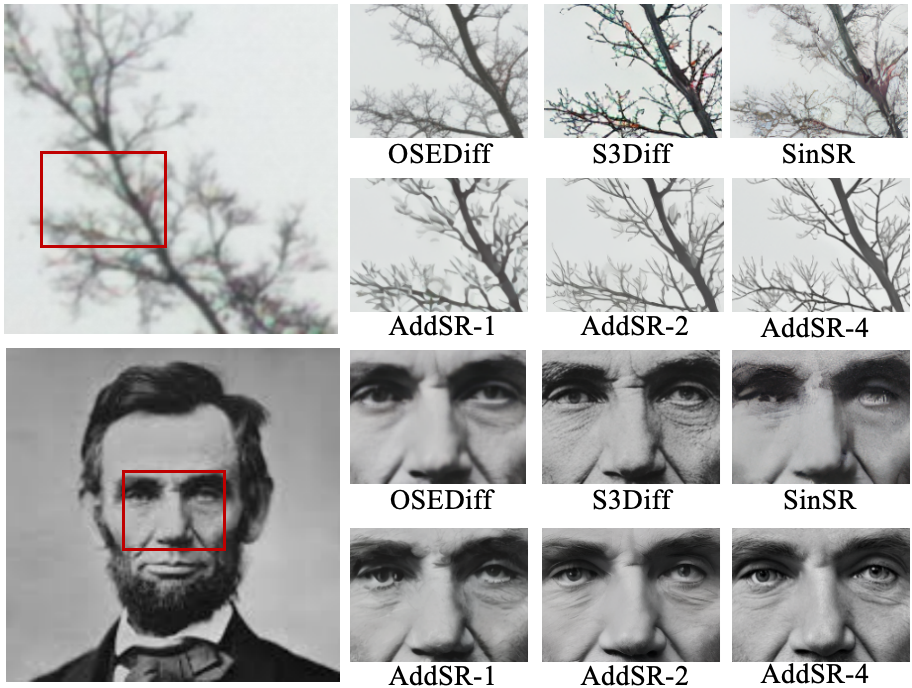}
    \vspace{-3mm}
    \caption{Qualitative comparison with other efficient methods. AddSR achieves comparable results within a single step and surpasses other efficient methods after 2 or 4 steps (Zoom in for details). }
    \label{fig:compare_efficient_method}
\end{figure}

\begin{table*}[ht]
 \newcommand{\tabincell}[2]{\begin{tabular}{@{}#1@{}}#2\end{tabular}}
 \begin{center}
  \caption{Complexity comparison of model complexity. The inference time of each method is calculated on 128 × 128 input images for 4x SR on an A100 GPU.}
  \label{metric_runtime}
\resizebox{0.85\textwidth}{!}{
  \begin{tabular}{c|cccccccc}
                \hline & \makecell[c]{StableSR} & \makecell[c]{DiffBIR} & \makecell[c]{SeeSR} & \makecell[c]{ResShift} & \makecell[c]{SinSR} & \makecell[c]{OSEDiff} & \makecell[c]{S3Diff} & \makecell[c]{\textbf{AddSR~(Ours)}}\\
                \hline
                \makecell[c]{Inference steps} & 200 & 50 & 50 & 15 & 1 & 1 & 1 & 1 / 4 \\
                Runtime [s] & 17.75 & 16.06 & 5.64 & 1.65 & 0.42 & 0.60 & 0.62 & 0.63 / 0.81 \\
                \hline
  \end{tabular}
}
 \end{center}
\end{table*}

\begin{table*}[h]
    \caption{Ablation studies on refined training process. The best results are marked in \textbf{bold}.}
    \label{tab:refined training process}
    \centering
    \resizebox{\textwidth}{!}{
    \begin{tabular}{c|cc|ccc|cccc}
       \hline
       \multirow{2}{*}{Exp} & \multirow{2}{*}{\makecell[c]{Condi\\Image}}  & \multirow{2}{*}{RAM} & \multicolumn{3}{c|}{RealLR200} & \multicolumn{4}{c}{DrealSR}\\
       ~ & ~ & ~ & MANIQA$^*\uparrow$ & MUSIQ$^*\uparrow$ & CLIPIQA$^*\uparrow$ & MANIQA$^*\uparrow$ & MUSIQ$^*\uparrow$ & CLIPIQA$^*\uparrow$ & PSNR$\uparrow$\\ \hline
       (1) & \ding{55}  & \ding{55}  & 0.5623 & 70.75 & 0.7431 & 0.5331 & 64.55 & 0.7285 & \textbf{26.96} \\
       (2) & \checkmark & \ding{55}  & 0.6092 & 71.76 & 0.7660 & 0.5372 & 62.67 & 0.6997 & 26.78 \\
       (3) & \ding{55}  & \checkmark & 0.5772 & 71.33 & 0.7549 & 0.5433 & 65.09 & 0.7087 & 26.87 \\
     \name & \checkmark & \checkmark & \textbf{0.6182} & \textbf{72.62} & \textbf{0.7724} & \textbf{0.6034} & \textbf{68.16} & \textbf{0.7381} & 26.09 \\ \hline
    \end{tabular}}
\end{table*}

\subsection{Comparison with Other Efficient Methods} 
\label{efficient method comparison}
We conduct a quantitative comparison with several other efficient methods \cite{zhang2024degradation, wang2024sinsr}, along with the Turbo versions of SeeSR \cite{wu2023seesr} and StableSR \cite{StableSR}, as shown in Tab.~\ref{tab:turbo}. 
AddSR-1 achieves comparable performance in MANIQA and MUSIQ metrics on RealLR200, as well as in PSNR, SSIM and MANIQA metrics on DrealSR. The primary reason AddSR-1 does not outperform other efficient methods is that those methods are specifically trained for one-step inference, which gives them an advantage in one-step setting. In contrast, AddSR offers greater flexibility, allowing for the application of multiple inference steps. As shown in Tab. \ref{tab:turbo}, AddSR-2 outperforms other efficient methods in the MANIQA and MUSIQ metrics on RealLR200 and DrealSR datasets. Furthermore, AddSR-4 achieves the best results on both MANIQA and MUSIQ metrics.

We also present a qualitative comparison in Figure \ref{fig:compare_efficient_method}. The visual results of AddSR-1 are comparable to those of other efficient methods. Thanks to the flexibility of AddSR, we can perform multiple inference steps. AddSR-2 and AddSR-4 produce clearer and more realistic results compared to other efficient methods.

\subsection{Complexity Analysis}
We compare diffusion-based methods in terms of the number of inference steps and runtime in Tab.~\ref{metric_runtime}. The runtime is calculated on an A100 GPU. From this comparison, we can make the following observations:
1) SinSR has the fastest runtime among other one-step methods, as it is not a SD-based method and has fewer total network parameters.
2) Among SD-based methods, OSEDiff, S3Diff and AddSR have similar runtimes in a single inference step, as their network structures are similar.

\subsection{Ablation Study}
\label{ablation study}

\noindent \textbf{Effectiveness of Refined Training Process.}
To enrich the information provided by the teacher model, we refine the training process by substituting the LR image with HR image as inputs of ControlNet, RAM and CLIP.
Since SeeSR is adopted as the baseline, we also replace the LR image of its RAM input with HR image. 
The quantitative results are shown in Tab.~\ref{tab:refined training process}. 
With the supervision from HR input, the perception quality of restored images becomes better.

\begin{table*}[]
    \centering
    \caption{Comparing the exponential form of timestep-adaptive ADD across different hyper-parameters.}
    \label{tab:exponential form}
    \resizebox{\textwidth}{!}{
    \begin{tabular}{c c|cc|cc|cc|cc}
    \hline
       \multirow{3}{*}{$\mu$} & \multirow{3}{*}{$\nu$} & \multicolumn{8}{c}{RealSR}\\ 
       ~ & ~ & \multicolumn{2}{c}{$1$ step} & \multicolumn{2}{c}{$2$ step} &\multicolumn{2}{c}{$3$ step} & \multicolumn{2}{c}{$4$ step}\\ 
       ~ & ~ & MANIQA$\uparrow$ & PSNR$\uparrow$ & MANIQA$\uparrow$ & PSNR$\uparrow$& MANIQA$\uparrow$ & PSNR$\uparrow$& MANIQA$\uparrow$ & PSNR$\uparrow$ \\ \hline
       \multirow{4}{*}{$0.5$} & 1.3 & 0.4202 & 24.11 & 0.6263 & 23.10 & 0.6427 & 22.36 & 0.6453 & 22.33 \\
       ~ & 1.7 & 0.3986 & 24.18 & 0.6110 & 24.01 & 0.6195 & 23.44 & 0.6307 & 23.40 \\ 
       ~ & \cellcolor{lightgray}2.1 & \cellcolor{lightgray}0.4189 & \cellcolor{lightgray}24.22 & \cellcolor{lightgray}0.6339 & \cellcolor{lightgray}23.29 & \cellcolor{lightgray}0.6496 & \cellcolor{lightgray}22.76 & \cellcolor{lightgray}0.6597 & \cellcolor{lightgray}22.73 \\
       ~ & 2.5 & 0.4207 & 24.48 & 0.5939 & 23.92 & 0.6081 & 23.33 & 0.6197 & 23.29 \\ \hline\hline
       0.7 & \multirow{2}{*}{$2.1$} & 0.3821 & 24.90 & 0.5971 & 24.03 & 0.6078 & 23.39 & 0.6221 & 23.36 \\
       0.9 & ~ & 0.4095 & 23.16 & 0.6052 & 22.97 & 0.6062 & 22.88 & 0.6244 & 22.68 \\ \hline
    \end{tabular}}
\end{table*}

\begin{table*}[t]
    \centering
    \caption{Comparing the linear form of timestep-adaptive ADD across different hyper-parameters.}
    \label{tab:linear form}
    \resizebox{0.92\textwidth}{!}{
    \begin{tabular}{c c|cc|cc|cc|cc}
    \hline
       \multirow{3}{*}{$\gamma$} & \multirow{3}{*}{$\kappa$} & \multicolumn{8}{c}{RealSR}\\ 
       ~ & ~ & \multicolumn{2}{c}{$1$ step} & \multicolumn{2}{c}{$2$ step} &\multicolumn{2}{c}{$3$ step} & \multicolumn{2}{c}{$4$ step}\\ 
       ~ & ~ & MANIQA$\uparrow$ & PSNR$\uparrow$ & MANIQA$\uparrow$ & PSNR$\uparrow$& MANIQA$\uparrow$ & PSNR$\uparrow$& MANIQA$\uparrow$ & PSNR$\uparrow$ \\ \hline
       0.1 & 0.7 & 0.3908 & 24.28 & 0.5849 & 23.64 & 0.6133 & 23.00 & 0.6172 & 22.98 \\ \hline\hline
       \multirow{3}{*}{$0.2$} & 0.3 & 0.4574 & 23.69 & 0.6294 & 23.06 & 0.6465 & 22.48 & 0.6480 & 22.41 \\
       ~ & 0.5 & 0.4338 & 23.87 & 0.6237 & 23.17 & 0.6407 & 22.55 & 0.6443 & 22.52 \\
       ~ & 0.7 & 0.4225 & 24.02 & 0.6064 & 23.24 & 0.6313 & 22.60 & 0.6352 & 22.59 \\ \hline\hline
       \multirow{3}{*}{$0.4$} & 0.1 & 0.4027 & 24.41 & 0.6077 & 23.25 & 0.6157 & 22.51 & 0.6215 & 22.45 \\
       ~ & 0.3 & 0.4045 & 24.39 & 0.5981 & 23.30 & 0.6124 & 22.58 & 0.6202 & 22.51 \\
       ~ & \cellcolor{lightgray}0.5 & \cellcolor{lightgray}0.4152 & \cellcolor{lightgray}24.73 & \cellcolor{lightgray}0.6261 & \cellcolor{lightgray}23.33 & \cellcolor{lightgray}0.6495 & \cellcolor{lightgray}22.65 & \cellcolor{lightgray}0.6507 & \cellcolor{lightgray}22.62 \\ \hline\hline
       \multirow{2}{*}{$0.6$} & 0.1 & 0.4156 & 24.48 & 0.6175 & 23.44 & 0.6261 & 22.82 & 0.6328 & 22.79 \\
       ~ & 0.3 & 0.3926 & 24.90 & 0.5905 & 23.93 & 0.5857 & 23.51 & 0.5981 & 23.42 \\ \hline\hline
       0.8 & 0.1 & 0.3887 & 24.92 & 0.5943 & 23.84 & 0.6038 & 23.36 & 0.6130 & 23.28 \\ \hline
    \end{tabular}}
\end{table*}

\noindent \textbf{Comparison of Timestep-Adaptive ADD Forms.}
To determine the optimal settings for timestep-adaptive ADD, we conduct experiments on its different forms: exponential and linear. 
Specifically, the exponential form is defined as Eq.~\ref{eq:exponential}, while the linear form is defined as follows:
\begin{equation}
   d(s, t) = (\prod_{i=0}^{t}(1-\beta_t))^\frac{1}{2} \times (\gamma \cdot p(s) + \kappa)
   \label{eq:linear}
\end{equation}
\noindent where the hyper-parameter $\kappa$ sets the initial weighting ratio, while $\gamma$ controls the increase of distillation loss over student timesteps.
The quantitative results of the exponential and linear forms under various settings are listed in Tab.~\ref{tab:exponential form} and Tab.~\ref{tab:linear form}, respectively. 
The best settings for different forms in the tables are highlighted with a gray background. 
From these tables, we can draw the following conclusions: 
(1) The best results of the exponential form are better than those of the linear form. Therefore, we use Eq.~(\ref{eq:exponential}) as the distillation loss function. Moreover, when $\mu=0.5$ and $\nu=2.1$, we achieves the best perceptual quality while maintaining good fidelity, so we use this setting for Eq.~(\ref{eq:exponential}). (2) Increasing the hyper-parameters that control the distillation loss ratio (i.e., $\nu$ and $\gamma$) typically results in higher fidelity. For instance, when we fix $\mu$ to 0.5 and increase $\nu$, the overall trend in 4 step shows a decrease in perception quality and an improvement in fidelity. Consequently, we can achieve a perception-distortion trade-off by adjusting $\nu$.

\begin{table*}[t!]
    \centering
    \caption{Ablation studies on PSR and TA-ADD.}
    \label{tab:PSR_TA_ADD}
    \resizebox{\textwidth}{!}{
    \begin{tabular}{cc|ccc|cccc}
        \toprule
        \multirow{2}{*}{Methods}  & \multirow{2}{*}{Time[s]} & \multicolumn{3}{c}{RealLR200} & \multicolumn{4}{c}{DrealSR} \\
        ~ & ~ & MANIQA$^*\uparrow$ & MUSIQ$^*\uparrow$ & CLIPIQA$^*\uparrow$ & MANIQA$^*\uparrow$ & MUSIQ$^*\uparrow$ & CLIPIQA$^*\uparrow$ & PSNR$\uparrow$ \\ \midrule
        w/o PSR & 0.63$\sim$0.77 & 0.5863 & 72.07 & 0.7631 & 0.5752 & 66.67 & 0.7243 & 26.52 \\
        w/o TA-ADD & 0.63$\sim$0.81 & 0.6058 & 72.19 & 0.7630 & 0.5898 & 67.58 & 0.7042 & 25.77 \\
        \name~ & 0.63$\sim$0.81 & \textbf{0.6182} & \textbf{72.62} & \textbf{0.7724} & \textbf{0.6034} & \textbf{68.16} & \textbf{0.7381} & \textbf{26.09} \\ \bottomrule
    \end{tabular}}
\end{table*}

\noindent \textbf{Effectiveness of TA-ADD on Balancing Perception and Fidelity.}
The proposed TA-ADD aims to balance perception and fidelity quality of restored images. 
The quantitative results are shown in Tab.~\ref{tab:PSR_TA_ADD}. 
Despite we increase the weight of the distillation loss in the later inference steps, the perceptual quality still improves. 
This could be attributed to the initial steps producing sufficiently high perception quality images, which offer more informative cues when combined with PSR. 
Consequently, the later inference steps can achieve high perceptual quality. 

In addition, we can adjust the hyperparameters in TA-ADD and the number of inference steps to achieve competitive PSNR and SSIM results while excelling in perceptual quality. We primarily compare the leading methods in terms of fidelity (GAN-based Real-ESRGAN) and perceptual quality (SeeSR). As shown in Tab.~\ref{tab:fidelity}, Our method remarkably enhances perceptual quality compared to SeeSR while also offering better fidelity. When compared to Real-ESRGAN, our method shows a substantial improvement in perceptual quality while maintaining comparable fidelity. This indicates that TA-ADD effectively navigates the perception-fidelity trade-off. Specifically, we made the following adjustments: 1) larger values for $\mu$ and $\nu$ ($\mu$=0.7, $\nu$=2.1) in TA-ADD during training, and 2) fewer inference steps (2 steps) to achieve high-fidelity results.

\begin{figure}[t!]
    \centering
    \includegraphics[width=1\linewidth]{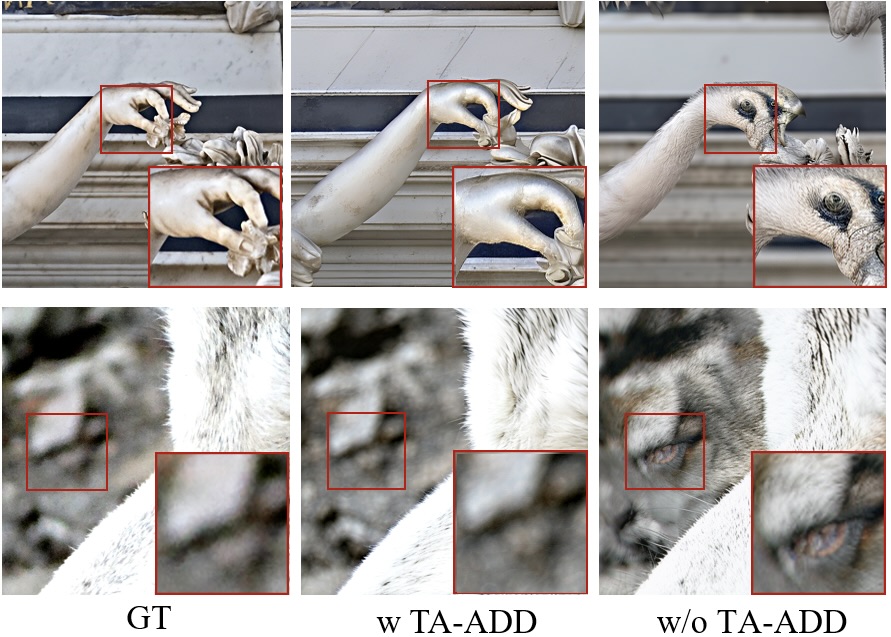}\
    \caption{Visual comparison of TA-ADD.}
    \label{fig:ta loss}
\end{figure}

The visual results are shown in Fig.~\ref{fig:ta loss}-Right. 
For the upper 3 images, the content is a statue.
However, without TA-ADD, the model hallucinates its hand as a bird. 
For the bottom 3 images, the original background is rock. 
Again, without utilizing TA-ADD,~\name~might hallucinate the background as an eye of a wolf. 
Conversely, with the help of TA-ADD, the restored images can generate more consistent contents with GTs. 
TA-ADD constrains the model from excessively leveraging its generative capabilities, thereby preserving more information in the image content, aligning closely with the GTs. 
Specifically, using TA-ADD, texture of the statue's hand in the upper image remains unchanged, and the background of the bottom image retains the rock with out-of-focus appearance.

\noindent \textbf{Effectiveness of PSR.}
As shown in Tab.~\ref{tab:PSR_TA_ADD}, incorporating PSR significantly enhances perceptual quality with minimal computational cost.
All of the three perception metrics, including MANIQA, MUSIQ and CLIPIQA, are improved on the two popular real-world datasets, namely RealLR200 and DrealSR.
Moreover, we use a additional degradation removal model to replace the PSR to demonstrate the effectiveness of PSR. Following StableSR \cite{StableSR} and DiffBIR \cite{DiffBIR}, we use Real-ESRGAN \cite{realesrgan} or SwinIR \cite{swinir} to replace the PSR. The quantitative results are shown in Table \ref{tab:degrdation removal}. We find that using Real-ESRGAN or SwinIR achieve comparable results to those obtained using our proposed PSR. However, the use of Real-ESRGAN or SwinIR compromises efficiency, as it requires an additional degradation removal model.

We also attempt to apply PSR to other SD-based methods without fine-tuning. However, we find that directly applying PSR to existing SD-based methods often results in collapsed outputs. The main reasons for this phenomenon can be summarized as follows:
1) Other SD-based methods typically require dozens of steps, which causes the initially predicted HR image to deviate significantly from the ground truth (GT). As the denoising process progresses, these errors accumulate, ultimately leading to significant deviations in the final results.
2) Due to the lack of fine-tuning, the distribution of the ControlNet input predicted HR image differs significantly from the LR images used during pre-training. This discrepancy prevents the model from effectively handling this condition, ultimately leading to the collapse of the final outputs.

\begin{table*}[t!]
    \caption{Quantitative comparison of different settings for TA-ADD on synthetic degraded DIV2K. The best results are \textbf{bold}, and the second best results are \underline{underlined}.}
    \centering
    \label{tab:fidelity}
    \resizebox{0.7\textwidth}{!}{
    \begin{tabular}{ccccc}
    \hline
        Metrics & Real-ESRGAN & SeeSR & Ours-perception (default) & Ours-fidelity \\ \hline
        MANIQA$\uparrow$ & 0.3374 & 0.5266 & \textbf{0.6335} & \underline{0.5759} \\
        CLIPIQ$\uparrow$ & 0.5047 & 0.7180 & \textbf{0.7703} & \underline{0.7357} \\
        PSNR$\uparrow$ & \textbf{22.70} & 21.86 & 21.45 & \underline{21.92} \\
        SSIM$\uparrow$ & \textbf{0.5935} & 0.5474 & 0.5210 & \underline{0.5481} \\ \hline
    \end{tabular}}
\end{table*}

\begin{table*}[t!]
    \caption{Comparison on different degradation removal methods.}
    \centering
    \label{tab:degrdation removal}
    \resizebox{\textwidth}{!}{
    \begin{tabular}{c|ccc|cccc}
    \hline
        \multirow{2}{*}{Methods} & \multicolumn{3}{c}{RealLR200} & \multicolumn{4}{c}{DrealSR}\\
        ~ & MANIQA$\uparrow$ & MUSIQ$\uparrow$ & CLIPIQA$\uparrow$ & PSNR$\uparrow$ & SSIM$\uparrow$ & MANIQA$\uparrow$ & CLIPIQA$\uparrow$\\ \hline
        w Real-ESRGAN & 0.5955 & 72.33 & 0.7692 & 26.03 & 0.6998 & 0.6012 & 0.7212 \\
        w SwinIR & 0.5956 & 72.32 & 0.7679 & \textbf{26.21} & \textbf{0.7112} & 0.6026 & 0.7324 \\
        w PSR & \textbf{0.6182} & \textbf{72.62} & \textbf{0.7724} & 26.09 & 0.7036 & \textbf{0.6034} & \textbf{0.7381} \\ \hline
    \end{tabular}}
\end{table*}

\section{Inference Strategies}
In this section, we discuss the impact of different inference strategies on AddSR. We mainly consider two aspects: 1) the blended PSR, as discussed in Sec. \ref{infer strategy: psr}; and 2) the impact of different positive prompts, as explored in Sec. \ref{infer strategy: pos. prompt}. 

\begin{figure*}[t!]
    \centering
    \includegraphics[width=1\linewidth]{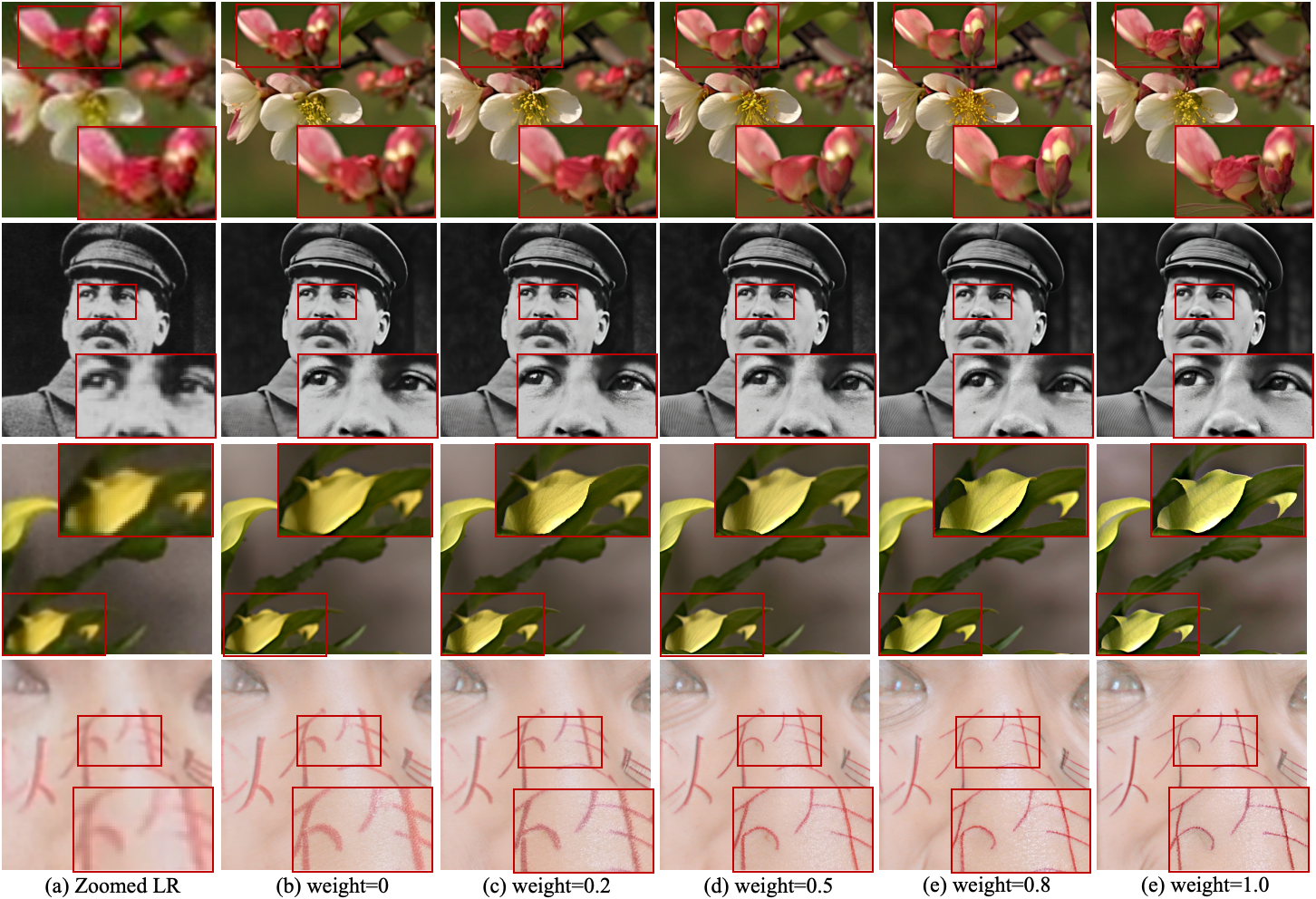}
    \vspace{-3mm}
    \caption{Qualitative results under different PSR weights on RealLR200 and DrealSR datasets (Zoom in for details).}
    \label{fig:psr_weight}
\end{figure*}

\subsection{Blended Prediction-\\based Self-Refinement}
\label{infer strategy: psr}

\begin{table*}[t!]
    \caption{Quantitative results under different blending ratio of PSR. We set the ratio to 1 as the default setting.}
    \centering
    \label{tab:blending ratio}
    \resizebox{0.9\textwidth}{!}{
    \begin{tabular}{c|ccc|cccc}
    \hline
        \multirow{2}{*}{Ratio} & \multicolumn{3}{c}{RealLR200} & \multicolumn{4}{c}{DrealSR}\\
        ~ & MANIQA$\uparrow$ & MUSIQ$\uparrow$ & CLIPIQA$\uparrow$ & PSNR$\uparrow$ & SSIM$\uparrow$ & MANIQA$\uparrow$ & CLIPIQA$\uparrow$\\ \hline
        0 & 0.5863 & 72.07 & 0.7631 & 26.52 & 0.7342 & 0.5752 & 0.7243 \\
        0.2 & 0.5892 & 72.23 & 0.7633 & 26.43 & 0.7289 & 0.5801 & 0.7313 \\ 
        0.5 & 0.5939 & 72.37 & 0.7699 & 26.27 & 0.7157 & 0.5962 & 0.7324 \\
        0.8 & 0.6016 & 72.49 & 0.7707 & 26.19 & 0.7099 & 0.6012 & 0.7365 \\
        1.0 & 0.6182 & 72.62 & 0.7724 & 26.09 & 0.7036 & 0.6034 & 0.7381 \\
        \hline
    \end{tabular}}
\end{table*}

To mitigate error accumulation caused by the predicted HR image, we can use a mixture of the LR image and the predicted HR image to control the model output, rather than relying solely on the predicted HR image. This approach can be formulated as follows:
\begin{equation}
   x_{mix} = r \hat{x}_0^i + (1-r) x_{LR}
   \label{eq5}
\end{equation}
where $x_{mix}$ denotes mixture of $\hat{x}_0$ and $x_{LR}$, $r$ is blending ratio.
As shown in Tab. \ref{tab:blending ratio}, a ratio of 0 indicates that only the LR image is used as the ControlNet input, while a ratio of 1 indicates that only the predicted HR image is used as the ControlNet input. Based on the results in the table, we can draw the following conclusions:
1) Increasing the blending ratio, \textit{i.e.,} increasing the proportion of the predicted HR image, enhances the perceptual quality. This is evidenced by the gradual improvement of the MANIQA, MUSIQ, and CLIPIQA metrics on both real-world datasets as the ratio increases. This improvement can be attributed to the high-frequency information in the predicted HR image, which enriches the details of the restored results.
2) Fidelity decreases as the blending ratio increases. Specifically, the PSNR and SSIM metrics on the DrealSR dataset decrease as the blending ratio grows. The primary reason for this is that, although the predicted HR image contains more high-frequency information, it may introduce details that do not exist in the LR image. Such errors accumulate during denoising, ultimately compromising fidelity. 

The visual results are shown in Fig. \ref{fig:psr_weight}. As the blending ratio increases, the results become clearer, as illustrated by the examples in the first and third rows. However, an excessively high ratio may also introduce errors. For instance, in the second and fourth rows, while the results appear clearer and more realistic, certain details deviate from the original LR images, such as the thinner fonts in the fourth-row example.

\begin{table*}[htbp]
  \caption{Comparison of different prompts and guidance strengths. Note that $s=0$ is equivalent to using negative prompts w/o guidance. Positive prompts are [1] ``\texttt{clean, high-resolution, 8k, extremely detailed, best quality, sharp}'', and [2] ``\texttt{Good photo}''. The Second row is the default settings for AddSR.}
  \label{tab:prompt}
    \resizebox{\textwidth}{!}{
    \begin{tabular}{c|ccc|cccc}
    \hline
        \multirow{2}{*}{Pos. Prompts} & \multicolumn{3}{c}{RealLR200} & \multicolumn{4}{c}{DrealSR}\\
        ~ & MANIQA$\uparrow$ & MUSIQ$\uparrow$ & CLIPIQA$\uparrow$ & PSNR$\uparrow$ & SSIM$\uparrow$ & MANIQA$\uparrow$ & CLIPIQA$\uparrow$\\ \hline
        [] & 0.5895 & 72.56 & 0.7674 & 26.27 & 0.7139 & 0.5911 & 0.7311 \\ 
        $[1]$ & 0.6182 & 72.62 & 0.7724 & 26.09 & 0.7036 & 0.6034 & 0.7381 \\ 
        $[2]$ & 0.5909 & 72.56 & 0.7666 & 25.95 & 0.6998 & 0.5932 & 0.7345 \\
        \hline
    \end{tabular}}
\end{table*}

\begin{figure}
    \centering
    \includegraphics[width=0.9\linewidth]{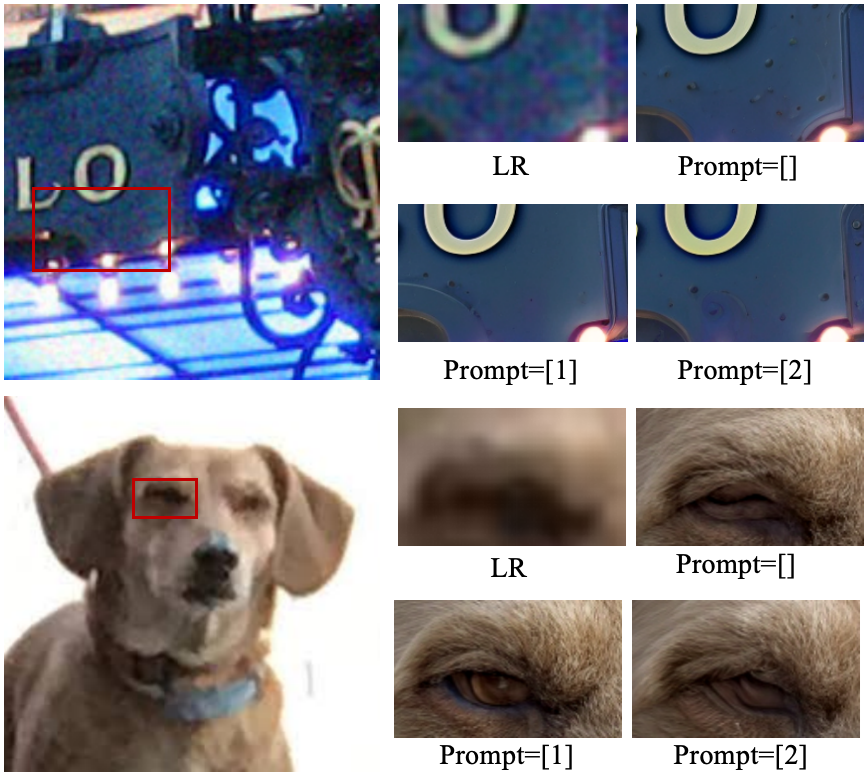}
    \caption{Qualitative results under different prompts (Zoom in for details).}
    \label{fig:prompt}
\end{figure}

\begin{figure*}[]
    \centering
    \includegraphics[width=0.85\linewidth]{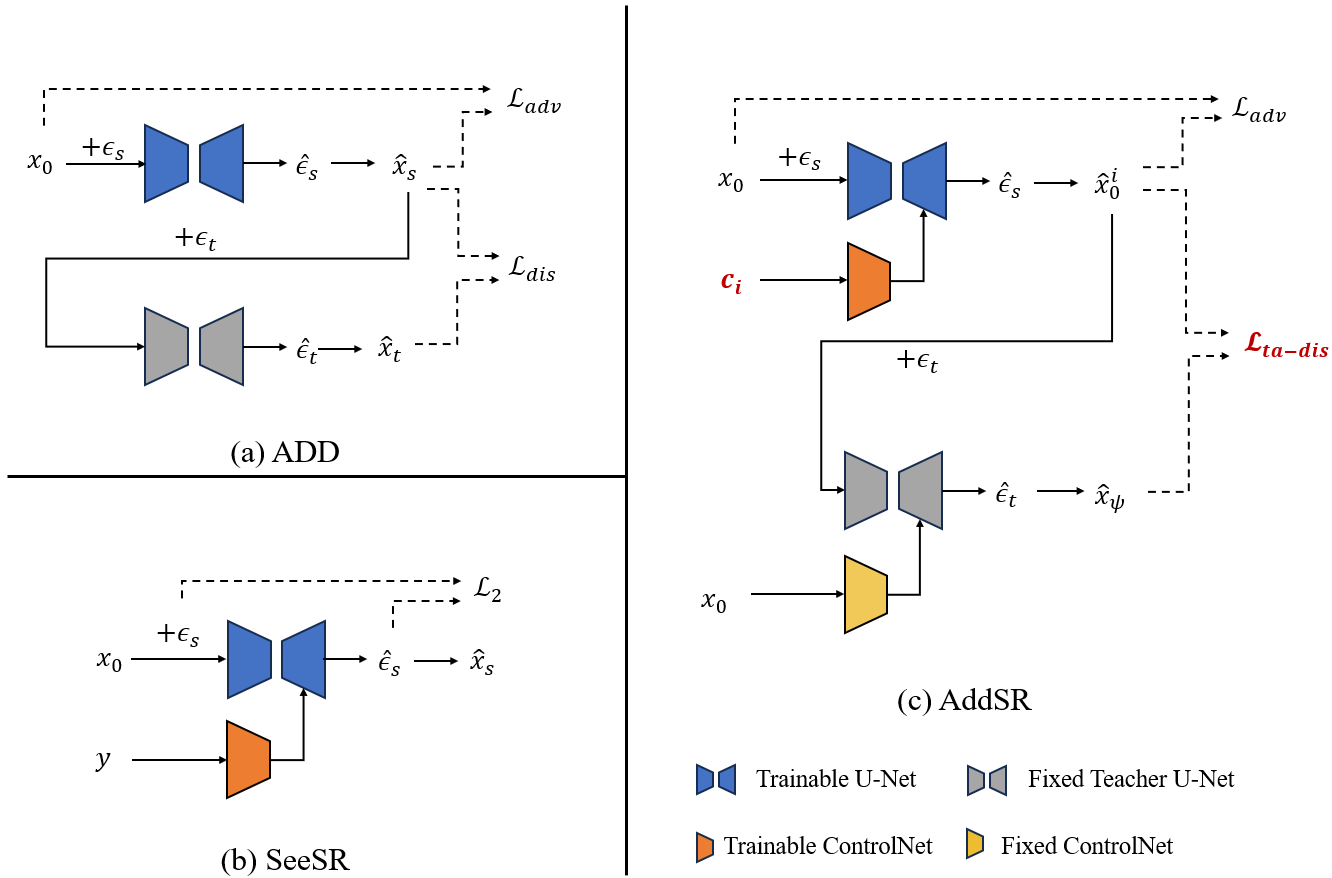}
    \caption{Comparisons on architecture diagram among ADD, SeeSR and~\name. 
    $x_0$ and y denote HR and LR images, respectively. 
    $\hat{x}_0^i$ and $\hat{x}_\phi$ denote the predicted $x_0$ from the timesteps $s$ and $t$, respectively. 
    $\epsilon_s$, $\epsilon_t$, $\hat{\epsilon}_s$ and $\hat{\epsilon}_s$ stand for added and predicted noise in timesteps $s$ and $t$, respectively. 
    $\mathcal{L}_{ta-dis}$ is the timestep-adaptive distillation loss.}
    \label{fig:comparsion}
\end{figure*}

\subsection{Positive Prompts Impact}
\label{infer strategy: pos. prompt}
Since AddSR does not utilize classifier-free guidance \cite{ho2022classifier} during inference, our analysis primarily focuses on the impact of positive prompts. Following StableSR \cite{StableSR}, we compare three settings: 1) no positive prompt, denoted as $[]$; 2) ``\texttt{clean, high-resolution, 8k, extremely detailed, best quality, sharp}'', denoted as $[1]$; and 3) ``\texttt{Good photo}'', denoted as $[2]$. As shown in Tab.~\ref{tab:prompt}, using positive prompt [1] slightly increases the perceptual quality metrics but compromises result fidelity, similar to the effect of using positive prompt [2]. 
The primary reason is that with a positive prompt, the restored results generate clearer details, improving perceptual quality metrics but deviating from the original LR image, reducing fidelity. As shown in Fig.~\ref{fig:prompt}, using positive prompt [1] helps the SR model better remove degradation (e.g., the first example) and generate more accurate structures, such as the dog's eyes in the second example.

\section{Discussion}
\label{discussion}

In this section, we provide a comparison among ADD, SeeSR, and our proposed~\name. Their architecture diagrams are depicted in Fig.~\ref{fig:comparsion}. 
Firstly, the distinctions between ADD and~\name~primarily lie in two aspects:
$1$) \textit{Introduction of ControlNet}: ADD is originally developed for text-to-image task, which typically only takes text as input. 
In contrast,~\name~is an image-to-image model that requires the additional ControlNet to receive information from the LR image.
$2$) \textit{Perception-distortion Trade-off}: ADD aims to generate photo-realistic images from texts. 
However, introducing ADD into blind SR brings the perception-distortion imbalance issue (please refer to Sec.~$3.4$ in our submission), which is addressed by our proposed timestep-adaptive ADD in~\name.

Secondly, the key differences between SeeSR and~\name~are:
$1$) \textit{Introduction of Distillation}: SeeSR is trained based on vanilla SD model that needs $50$ inference steps, while~\name~utilizes a teacher model to distill an efficient student model to achieve just $1$$\sim$$4$ steps.
$2$) \textit{High-frequency Information}: SeeSR uses the LR image $y$ as the input of the ControlNet. In contrast,~\name~on one hand adopts the HR image $x_0$ as the input of the teacher model's ControlNet to supply the high-frequency signals since the teacher model is not required during inference. 
On the other hand,~\name~proposes a novel prediction-based self-refinement (PSR) to further provide high-frequency information by replacing the LR image with the predicted image as the input of the student model's ControlNet.
Therefore,~\name~has the ability to generate results with more realistic details.

\section{Conclusion}
\label{sec:conclusion}
We propose~\name, an effective and efficient model based on Stable Diffusion prior for blind super-resolution. 
To address the perception-distortion imbalance issue introduced by the original ADD, we introduce timestep-adaptive ADD, which assigns distinct weights to GAN loss and distillation loss across different student timesteps.
In contrast to current SD-based BSR approaches that either use LR images to regulate each inference step's output or rely on additional modules to pre-clean LR images as conditions,~\name~substitutes the LR image with the HR image estimated in the preceding step. This substitution provides more high-frequency information, allowing for restored results with enhanced textures and edges, while maintaining efficiency.
Additionally, we use the HR image as the controlling signal for the teacher model, enabling it to provide better supervision.
Extensive experiments demonstrate that~\name~generates superior results within $1$$\sim$$4$ steps in various degradation scenarios and real-world low-quality images.

\noindent \textbf{Limitations.}
\label{sec:limitations}
Although the inference speed of our~\name~surpasses all of the existing SD-based methods remarkably, there still exists a gap between~\name~and GAN-based methods. 
The primary factor is that~\name~is built upon SD and ControlNet, which, due to its substantial model parameters and intricate network structure, noticeably hinders the inference time. 
In the future, we plan to explore a more streamlined network architecture to boost overall efficiency.

\bibliography{sn-bibliography}

\end{document}